\documentclass{article} %
\usepackage{iclr2025_conference,times}

\usepackage{amsmath,amsfonts,bm}

\def\eqref#1{equation~\ref{#1}}

\def\1{\bm{1}}

\DeclareMathAlphabet{\mathsfit}{\encodingdefault}{\sfdefault}{m}{sl}
\SetMathAlphabet{\mathsfit}{bold}{\encodingdefault}{\sfdefault}{bx}{n}

\DeclareMathOperator*{\argmax}{arg\,max}

\usepackage[utf8]{inputenc} %
\usepackage[T1]{fontenc}    %
\usepackage{hyperref}       %
\usepackage{url}            %
\usepackage{booktabs}       %
\usepackage{amsfonts}       %
\usepackage{nicefrac}       %
\usepackage{microtype}      %
\usepackage{xcolor}         %

\usepackage{xspace}
\usepackage{amsmath}
\usepackage{multirow}
\usepackage{graphicx}
\usepackage{amssymb}
\usepackage{mathtools}
\usepackage{amsthm}

\usepackage{caption}
\usepackage{subcaption}
\usepackage{tcolorbox}

\usepackage{algorithm}
\usepackage{algorithmic}

\usepackage{adjustbox}
\usepackage[normalem]{ulem}

\usepackage{tikz,tkz-graph}
\usetikzlibrary{
  graphs,
  graphs.standard,
  positioning,chains,fit,shapes,calc,
  patterns
}
\usepackage{graphicx,subcaption}
\usepackage{wrapfig}

\usepackage{amsmath} %
\usepackage{tikz}
\usepackage{etoolbox} %
\usepackage{listofitems} %
\usetikzlibrary{arrows.meta} %
\usepackage[outline]{contour} %
\contourlength{1.4pt}

\tikzset{>=latex} %
\usepackage{xcolor}
\colorlet{myred}{red!80!black}
\colorlet{myblue}{blue!80!black}
\colorlet{myyellow}{yellow!80!black}
\colorlet{mygreen}{green!60!black}
\colorlet{myorange}{orange!70!red!60!black}
\colorlet{mydarkred}{red!30!black}
\colorlet{mydarkblue}{blue!40!black}
\colorlet{mydarkgreen}{green!30!black}
\tikzstyle{node}=[very thick,circle,draw=myblue,minimum size=22,inner sep=0.5,outer sep=0.6]
\tikzstyle{node in}=[node,green!15!black,draw=mygreen,fill=mygreen!25]
\tikzstyle{node hidden}=[node,yellow!15!black,draw=myyellow,fill=myyellow!20]
\tikzstyle{node hidden2}=[node,blue!15!black,draw=myblue,fill=myblue!20]
\tikzstyle{node convol}=[node,orange!15!black,draw=brown,fill=brown!20]
\tikzstyle{node out}=[node,red!15!black,draw=myred,fill=myred!20]
\tikzstyle{connect}=[thick,mydarkblue] %
\tikzstyle{connect arrow}=[-{Latex[length=4,width=3.5]},thick,mydarkblue,shorten <=0.5,shorten >=1]
\tikzset{ %
  node 0/.style={node convol},
  node 1/.style={node in},
  node 2/.style={node hidden},
  node 3/.style={node out},
}
\def\nstyle{int(\lay<\Nnodlen?min(2,\lay):3)} %
\usepackage[capitalize,noabbrev]{cleveref}

\theoremstyle{plain}

\theoremstyle{definition}

\theoremstyle{remark}

\usepackage[textsize=tiny]{todonotes}

\newcommand{\name}{LICO\xspace}
\newcommand{\blue}[1]{\textcolor{blue}{\mathbf{#1}}}
\newcommand{\violet}[1]{\textcolor{violet}{\mathbf{#1}}}

\title{\name{}: Large Language Models for In-Context Molecular Optimization}

\author{Tung Nguyen \& Aditya Grover \\
Department of Computer Science \\
University of California, Los Angeles \\
\texttt{\{tungnd,adityag\}@cs.ucla.edu} \\
}

\iclrfinalcopy %
\begin{document}

\maketitle

\begin{abstract}
Optimizing black-box functions is a fundamental problem in science and engineering.
To solve this problem, many approaches learn a surrogate function that estimates the underlying objective from limited historical evaluations.
Large Language Models (LLMs), with their strong pattern-matching capabilities via pretraining on vast amounts of data, stand out as a potential candidate for surrogate modeling.
However, directly prompting a pretrained language model to produce predictions is not feasible in many scientific domains due to the scarcity of domain-specific data in the pretraining corpora and the challenges of articulating complex problems in natural language.
In this work, we introduce \name{}, a general-purpose model that extends arbitrary base LLMs for black-box optimization, with a particular application to the molecular domain.
To achieve this, we equip the language model with a separate embedding layer and prediction layer, and train the model to perform in-context predictions on a diverse set of functions defined over the domain.
Once trained, \name{} can generalize to unseen molecule properties simply via in-context prompting.
\name{} performs competitively on PMO, a challenging molecular optimization benchmark comprising $23$ objective functions, and achieves state-of-the-art performance on its low-budget version PMO-1K. 

\end{abstract}

\section{Introduction}
Black-box optimization (BBO) is the problem of optimizing an unknown, often complex objective function without direct access to its structure or derivatives. This problem is ubiquitous in many science and engineering fields, including material discovery~\citep{hamidieh2018data}, protein engineering~\citep{brookes2019conditioning,sarkisyan2016local,angermueller2020model}, molecular design~\citep{gaulton2012chembl}, mechanical design~\citep{berkenkamp2016safe,liao2019data}, and neural architecture search~\citep{zoph2016neural}. Typically, BBO involves an iterative process where each step constructs a surrogate model to approximate the objective function. This model then guides the selection of promising candidates for subsequent evaluation. The main challenge of this approach lies in learning an effective surrogate function that can accurately estimate the objective from limited historical data. 

In stark contrast, we have seen impressive generalization abilities of Large Language Models (LLMs)~\citep{brown2020language,achiam2023gpt,bubeck2023sparks,team2023gemini,touvron2023llama,touvron2023llama2,jiang2023mistral,jiang2024mixtral} for language-driven reasoning over many kinds of domains. By pretraining on Internet-scale data, LLMs have demonstrated exceptional pattern-matching abilities and generalization from limited observations in both natural language~\citep{brown2020language,kojima2022large,wei2022chain} and other domains~\citep{lu2022frozen,mirchandani2023large,gruver2023large}. This positions LLMs as a promising solution for enhancing surrogate modeling for BBO.
Some recent works have indeed shown great potential for using LLMs for solving optimization problems~\citep{yang2023large,chen2023evoprompting,zhang2023using,liularge}. The main idea behind these methods is to frame the optimization problem in natural language, and prompt the language model using previously collected observations to make predictions for new data points~\citep{liularge} or to propose better candidates~\citep{yang2023large,chen2023evoprompting,zhang2023using,ma2023eureka,nie2023importance,meyerson2023language,lehman2023evolution,bradley2024openelm,liu2023large}. However, this approach has several limitations. First, performing optimization in the text space requires the problem and solution to be expressed in natural language, thus limiting this approach to selected domains. Second, the scarcity of domain-relevant data in the text corpora used to train language models poses generalization challenges when using these models for general scientific domains such as molecular optimization. Therefore, existing works have only demonstrated the success of LLMs in neural architecture search~\citep{liularge,chen2023evoprompting,zhang2023using}, prompt optimization~\citep{yang2023large}, and code generation~\citep{ma2023eureka,lehman2023evolution}, corresponding to domains that are well-represented in the training dataset for common language models~\citep{brown2020language,touvron2023llama,jiang2023mistral}. Third, relying on verbose textual descriptions for the problem and its solution imposes practical constraints by inflating the context length and reducing the number of historical observations the model can effectively utilize.

In this work, we propose \textbf{L}arge Language Models for \textbf{I}n-\textbf{C}ontext \textbf{O}ptimization (\name{}), a general-purpose model that leverages LLMs for black-box optimization, with a particular application to the molecular domain. To generalize a language model to a new scientific domain unseen during pretraining, we equip the model with two embedding layers for embedding the previously collected molecules and their scores, and a prediction head to predict the score of unseen candidates. Intuitively, the embedding layers map the molecules and their scores to the same feature space already learned by the language model, allowing the model to perform in-context learning in this space instead of the raw text space.
Unlike previous methods, this approach is applicable to domains that may not be easily described in natural language such as molecular optimization. Moreover, avoiding verbose textual descriptions enables the model to condition on more historical observations, thus scaling better to harder problems that cannot be solved within a few steps.

We train the new layers together with the (frozen) LLM to perform in-context predictions on a family of functions. Specifically, for each function sampled from this family, we condition the model on a set of inputs and their corresponding evaluations, and task the model to predict the function value of the remaining data points. This task mimics surrogate modeling in BBO, where the surrogate model has to iteratively update its estimation of the underlying objective by conditioning on historical data. 
An ideal function family to train the model should be close to the target objective functions we want to optimize, but also be diverse enough to encourage generalization. Therefore, we propose to combine intrinsic functions and synthetically generated functions for training \name{}. Intrinsic functions are inherent properties of the input that are easy to evaluate. In molecular optimization, for example, intrinsic functions include molecular weight, the number of rings, or heavy atom count, which are obtained via simple computation on the molecule. These intrinsic functions are closely related to the actual objective functions we want to optimize such as bioactivities against a target disease. To facilitate generalization outside of the intrinsic functions, we additionally train \name{} on synthetic functions defined over the same target domain that are generated by Gaussian Processes. Our empirical evidence shows the importance of learning from both intrinsic and synthetic functions to the performance of the model on downstream tasks. Figure~\ref{fig:architecture} illustrates our approach.

After training, \name{} is capable of optimizing a wide range of molecular properties purely via in-context prompting. While the methodology of \name{} applies to general scientific domains, in this paper we focus on molecular optimization. This problem plays a pivotal role in advancing drug and material discovery. The complexity of molecular structures and the vastness of the chemical space present unique challenges to black-box optimization algorithms. Moreover, since molecule-relevant data is likely under-represented in the pretraining corpora of existing language models, molecular optimization is a good problem to test the performance and applicability of \name{}. 
We demonstrate the competitive performance of \name{} against the leading methods on Practical Molecular Optimization (PMO)~\citep{gao2022sample}, a challenging molecular optimization benchmark with $23$ objective functions. On PMO's low-budget setting, which we term PMO-1M, \name{} achieves the best performance and is the highest-ranked method in the benchmark.

\begin{figure*}
\begin{subfigure}{\textwidth}
\centering
\begin{adjustbox}{width=0.9\textwidth}
\begin{tikzpicture}

\def\xa{-3}  %
\def\ya{-3}  %
\def\yq{\ya + 0.5}  %
\def\xb{\xa+6}  %
\def\yb{\ya+3.35} %
\def\xc{\xb+2+1.5}
\def\xu{-6}
\def\yv{-1}
\def\yz{\yq-2.}

\node[rectangle, draw,rounded corners, minimum width=0.9cm,minimum height=0.4cm, fill=gray!20, align=center] at (\xu-0.2, \yb+0.5){};
\node[black] at (\xu+1.5, \yb+0.5) {Text embedding};

\node[rectangle, draw,rounded corners, minimum width=0.9cm,minimum height=0.4cm, fill=mygreen!20, align=center] at (\xu-0.2, \yb-0.2){};
\node[black] at (\xu+1.4, \yb-0.2) {$x$ embedding};

\node[rectangle, draw,rounded corners, minimum width=0.9cm,minimum height=0.4cm, fill=myred!20, align=center] at (\xu-0.2, \yb-0.9){};
\node[black] at (\xu+1.4, \yb-0.9) {$y$ embedding};

\node[rectangle, draw,rounded corners, minimum width=0.9cm,minimum height=0.4cm, fill=myyellow!20, align=center] at (\xu-0.2, \yb-1.6){};
\node[black] at (\xu+1.5, \yb-1.6) {Prediction layer};

\node[cylinder, draw, minimum size=1.5cm, shape border rotate=90, cylinder uses custom fill, cylinder body fill=mygreen!10, cylinder end fill = mygreen!40] (cyld1) at (\xu+0.2,\yz) {$\Tilde{\mathcal{F}}$};

\node[black] (xt0) at (\xa+1.7, \yz) {\large \texttt{prompt}};

\node[black] (xt1) at (\xa+3, \yz) {\Large \texttt{<x>}};
\node[node 1, minimum size=24] (x1) at (\xa+4.2, \yz) {$x_1$};
\node[black] (yt1) at (\xa+5.4, \yz) {\Large \texttt{<y>}};
\node[node 3, minimum size=24] (y1) at (\xa+6.6, \yz) {$y_1$};

\draw[thick, ->] (cyld1.east) to[out=0, in=-180]  node[below, align=center] {\textbf{Training}} (xt0.west);
\node[black] at (\xa-0.6, \yz+0.25) {\textbf{Semi-synthetic}};

\filldraw[black] (\xa+7.5,\yz) circle (2pt);
\filldraw[black] (\xa+8,\yz) circle (2pt);
\filldraw[black] (\xa+8.5,\yz) circle (2pt);

\node[black] (xt2) at (\xa+9.4, \yz) {\Large \texttt{<x>}};
\node[node 1, minimum size=24] (x2) at (\xa+10.6, \yz) {$x_n$};
\node[black] (yt2) at (\xa+11.8, \yz) {\Large \texttt{<y>}};
\node[node 3, minimum size=24] (y2) at (\xa+13, \yz) {$y_n$};

\node[rectangle, draw,rounded corners, minimum width=12.7cm,minimum height=1.2cm, fill=myblue!20, align=center] (tr) at (\xb+1.3, \ya+1.1){{\large Pretrained LLM} };

\node[rectangle, draw,rounded corners, minimum width=1cm,minimum height=0.6cm, fill=gray!20, align=center] (ext0) at (\xa+1.7, \yq-0.9){};
\node[rectangle, draw,rounded corners, minimum width=1cm,minimum height=0.6cm, fill=gray!20, align=center] (ext1) at (\xa+3, \yq-0.9){};
\node[rectangle, draw,rounded corners, minimum width=1cm,minimum height=0.6cm, fill=mygreen!20, align=center] (ex1) at (\xa+4.2, \yq-0.9){};
\node[rectangle, draw,rounded corners, minimum width=1cm,minimum height=0.6cm, fill=gray!20, align=center] (eyt1) at (\xa+5.4, \yq-0.9){};
\node[rectangle, draw,rounded corners, minimum width=1cm,minimum height=0.6cm, fill=myred!20, align=center] (ey1) at (\xa+6.6, \yq-0.9){};

\draw [->] (xt0) to  (xt0 |- ext0.south);
\draw [->] (xt1) to  (xt1 |- ext1.south);
\draw [->] (x1) to  (x1 |- ex1.south);
\draw [->] (yt1) to  (yt1 |- eyt1.south);
\draw [->] (y1) to  (y1 |- ey1.south);

\draw [->] (ext0) to  (ext0 |- tr.south);
\draw [->] (ext1) to  (ext1 |- tr.south);
\draw [->] (ex1) to  (ex1 |- tr.south);
\draw [->] (eyt1) to  (eyt1 |- tr.south);
\draw [->] (ey1) to  (ey1 |- tr.south);

\node[rectangle, draw,rounded corners, minimum width=1cm,minimum height=0.6cm, fill=gray!20, align=center] (ext2) at (\xa+9.4, \yq-0.9){};
\node[rectangle, draw,rounded corners, minimum width=1cm,minimum height=0.6cm, fill=mygreen!20, align=center] (ex2) at (\xa+10.6, \yq-0.9){};
\node[rectangle, draw,rounded corners, minimum width=1cm,minimum height=0.6cm, fill=gray!20, align=center] (eyt2) at (\xa+11.8, \yq-0.9){};
\node[rectangle, draw,rounded corners, minimum width=1cm,minimum height=0.6cm, fill=myred!20, align=center] (ey2) at (\xa+13, \yq-0.9){};

\draw [->] (xt2) to  (xt2 |- ext2.south);
\draw [->] (x2) to  (x2 |- ex2.south);
\draw [->] (yt2) to  (yt2 |- eyt2.south);
\draw [->] (y2) to  (y2 |- ey2.south);

\draw [->] (ext2) to  (ext2 |- tr.south);
\draw [->] (ex2) to  (ex2 |- tr.south);
\draw [->] (eyt2) to  (eyt2 |- tr.south);
\draw [->] (ey2) to  (ey2 |- tr.south);

\node[rectangle, draw,rounded corners, minimum width=1cm,minimum height=0.6cm, fill=myyellow!20, align=center] (py1) at (\xa+5.4, \yb-0.8){};
\node[rectangle, draw,rounded corners, minimum width=1cm,minimum height=0.6cm, fill=myyellow!20, align=center] (pym) at (\xa+11.8, \yb-0.8){};

\node[node 3, minimum size=24] (yhat1) at (\xa+5.4, \yb+0.4) {$\hat{y}_1$};
\node[node 3, minimum size=24] (yhatm) at (\xa+11.8, \yb+0.4) {$\hat{y}_n$};

\draw [<-] (py1) to (py1 |- tr.north);
\draw [<-] (pym) to (pym |- tr.north);

\draw [->] (py1) to  (py1 |- yhat1.south);
\draw [->] (pym) to  (pym |- yhatm.south);

\end{tikzpicture}
\end{adjustbox}
\end{subfigure}

\caption{Our proposed approach. We equip a pretrained LLM with an embedding layer for $x$, an embedding layer for $y$, and a prediction layer. We train the model on semi-synthetic data to predict $y$ given $x$ and previous $(x,y)$ pairs. We prepend each $x$ with a special token \texttt{<x>} and each $y$ with a special token \texttt{<y>} to guide in-context reasoning.}
\label{fig:architecture}
\end{figure*}
\section{Problem Statement}
Let $f: \mathcal{X} \rightarrow \mathbb{R}$ be a real-valued function that operates on a $d$-dimensional space $\mathcal{X} \subseteq \mathbb{R}^d$. In black-box optimization (BBO), the goal is to find the input $x^\star$ that maximizes $f$:
\begin{equation}
    x^\star \in \argmax_{x \in \mathcal{X}} f(x), \label{eq:optimization}
\end{equation}
where we do not have direct access to the structure or gradient information of $f$. In molecular optimization, $\mathcal{X}$ is the space of all possible molecules, and $f$ is a certain property of the molecule we want to optimize over, such as bioactivities against a disease. While $f$ is unknown, we often have access to an unlabeled dataset $\mathcal{D}_{\text{u}}$ that consists of molecules $x's$ without the corresponding function values $y's$. ZINC~\citep{sterling2015zinc} is such a dataset with thousands of unlabeled molecules.

To solve the optimization in~\eqref{eq:optimization}, we can query $f$ with a limited budget, since evaluation often involves expensive physical experiments. To overcome this challenge, a common BBO approach learns a surrogate model $f_\theta$ that approximates the objective $f$ from past observations \(\mathcal{D}_{\text{obs}} = \{(x_i,y_i)\}_{i=1}^n\), which starts empty and incrementally expands with new data points \((x, f(x))\) we query at each iteration.
Formally, a surrogate model represents a predictive distribution $p_\theta(y \mid x, \mathcal{D}_{\text{obs}})$ of the function value $y$ conditioned on the input $x$ and the evolving observed dataset $\mathcal{D}_{\text{obs}}$. 
The prediction of this surrogate guides the selection of candidates to balance exploration and exploitation. The newly selected points are added to $\mathcal{D}_{\text{obs}}$, and the process continues.

The success of this approach highly depends on the efficiency of the surrogate model $f_\theta$ in estimating $f$ from limited data in $\mathcal{D}_{\text{obs}}$ at each iteration.
This resembles few-shot prediction, 
a setting that Large Language Models (LLMs) have proven to excel in. By pretraining on vast Internet-scale data, LLMs can learn generalizable patterns from limited data, and are capable of adapting to multiple functions at test time simply via in-context prompting~\citep{brown2020language,mirchandani2023large,krishnamoorthy2022generative,krishnamoorthy2023diffusion}. A recent line of works~\citep{yang2023large,zhang2023using,chen2023evoprompting,liularge} has exploited this ability of LLMs for optimization, but they relied on natural language as the interface, thus lacking generality to scientific domains. In this work, we propose a more general and efficient approach to leveraging LLMs for black-box optimization.
\section{Related Work}

\textbf{LLMs for Optimization } Recent works have explored the use of LLMs for optimization. The general idea behind these works is to prompt the model with the textual description of the optimization problem and historical evaluations for few-shot reasoning.~\citet{yang2023large,liu2023large,zhang2023using,ma2023eureka} propose to prompt the language model to directly suggest better candidates to evaluate given the past inputs and their corresponding scores.~\citet{meyerson2023language,lehman2023evolution,bradley2024openelm} integrate LLMs with evolutionary algorithms, and prompt the model to perform crossover and mutation operations based on the population at each optimization step.~\citet{liularge} study the use of LLMs to enhance several components in Bayesian optimization, including warmstarting, surrogate modeling, and candidate generation. Optformer~\citep{chen2022towards} proposed to train an LLM specialized for in-context function prediction and optimization.

\textbf{LLMs for Molecular Optimization}
Recent works have proposed to leverage LLMs for molecular optimization via prompting~\citep{wang2024efficient,liu2023chatgpt,ramos2023bayesian,volker2024llms}, leveraging LLM embeddings~\citep{rankovic2023bochemian}, or finetuning on molecular corpora~\citep{guevorguian2024small,ye2023drugassist,fang2023domain,kristiadisober}. MOLLEO~\citep{wang2024efficient} and ChatDrug~\citep{liu2023chatgpt} are two prominent works in the first direction. MOLLEO proposed to prompt a pretrained LLM to perform crossover and mutation operations in a standard graph genetic algorithm, but its performance largely depends on the prompt format. Similarly, ChatDrug prompts a pretrained LLM for drug editing and requires a retrieval database of molecules to inject domain feedback into the LLM. BoChemian~\citep{rankovic2023bochemian} and~\citet{kristiadisober} studied the use of LLM embeddings for Bayesian molecular optimization, and~\citet{kristiadisober} additionally explored finetuning an LLM to serve as a surrogate for optimization. Chemlactica/Chemma~\citep{guevorguian2024small} and MOLGEN~\citep{fang2023domain} proposed to pretrain and/or finetune LLMs on molecule-related corpora to generate valid molecules, which can serve as a genetic algorithm in molecular optimization. DrugAssist~\citep{ye2023drugassist} creates the MolOpt-Instructions dataset that contains pairs of molecules and their property values to finetune a pretrained LLM that can iteratively propose better molecules after training.

The common approach in existing works has several inherent limitations. First, for general scientific domains, the input $x$ may not be easily described by natural language. Second, even when there is a textual description of the input, for instance, molecules can be represented by SMILES strings~\citep{weininger1988smiles}, existing prompt-based works require significant prompt optimization to achieve good performance, and the optimal prompt often varies between tasks.
Furthermore, from an engineering perspective, naively prompting a language model with verbose textual descriptions of the input $x$ results in an excessively long context, thus reducing the number of examples the model can condition on. 
For example, an LLM with a maximum context length of $4000$ can only utilize up to $100$ past observations, assuming the average length of each data point is $40$. This practically limits the scalability of this approach to harder problems that require more steps to solve.

\textbf{LLMs for Non-language Tasks } In addition to optimization, several works have studied the extension of pretrained LLMs to non-language domains with two main directions. The first direction considers problems that can be described in natural language, and prompts a pretrained LLM to solve the problem directly in the text space~\citep{mirchandani2023large,dinh2022lift,gruver2023large,liu2024conversational,sprueill2024chemreasoner}. The second direction tackles more general problems by learning separate encoders for the new domain and aligning it with the embedding space of the pretrained LLM~\citep{lu2022frozen,shen2023cross,tsimpoukelli2021multimodal,li2022pre}. Our work is closely related to the latter direction. However, as discussed in the following sections, while many of these works completely leave the word space, we find it beneficial to include language instruction while training the new modules.

\section{Method} \label{sec:method}
We introduce \name{}, a methodology for extending arbitrary base LLMs for surrogate modeling in black-box optimization. While the method applies to broad scientific domains, we choose molecular optimization to demonstrate \name{} in this paper.
We aim to develop a model capable of efficiently adapting to various objective functions after training. 
To achieve this, we propose a simple extension to existing LLMs and an unsupervised objective using semi-synthetic data to facilitate generalization.

\subsection{Model Architecture}
In black-box optimization, a surrogate model $f_\theta$ estimates the distribution of the function value $y$ given the input $x$ and past observations $\mathcal{D}_{\text{obs}} = \{(x_i,y_i)\}_{i=1}^n$ the model has collected until the optimization iteration $t$:
\begin{equation}
    p_\theta(y \mid x, x_1, y_1, x_2, y_2, \dots, x_n, y_n), \label{eq:estimation}
\end{equation}
where $x_i$ and $y_i = f(x_i)$ are drawn from an objective function $f$. 
Our goal is to explore LLMs to model $p_\theta$. As discussed earlier, we make no assumptions on the domain $\mathcal{X}$ to be expressed with natural language.
To extend a pretrained language model to an arbitrary new domain, we equip the model with $3$ new layers -- an embedding layer for the inputs $x's$, an embedding layer for the function values $y's$, and a prediction layer for predicting the unknown function value $y$. Learning separate embedding layers offers several benefits. First, the new embedding layers encode $x$ and $y$ to a shared hidden space obtained by the language model via pretraining, which enables the model to escape the raw text space and perform in-context reasoning in the hidden space instead. Moreover, by embedding each input $x$ to a single hidden vector instead of spanning it over several tokens, we effectively reduce the sequence length and thus allow the model to scale to more conditioning examples.

However, it is challenging for the model to perform this prediction task without any context information about the task. This is because, from the model point of view, embeddings of $x$ and $y$ do not mean anything more than some high-dimensional vectors.
In other words, the model does not know what task it should perform and what each token in the embedding sequence represents.
To address this issue, we prepend each sequence with a \texttt{task prompt} and prepend each input $x$ with a special token \texttt{<x>} and each function value $y$ with a special token \texttt{<y>}.
The \texttt{task prompt} instructs the model to perform the task, while the special tokens \texttt{<x>} and \texttt{<y>} inform the model of the position of each input $x$ and the corresponding function value $y$.
In other words, we use a language the model has mastered (natural language) to guide the learning of a new ``foreign language” (e.g., molecule).
In practice, the \texttt{task prompt} is \textit{``Each x is a molecule and each y is the property of the corresponding molecule. Predict y given x.”}, whereas \texttt{<x>} and \texttt{<y>} are two single characters \textit{``x''} and \textit{``y''}.
Finally, we apply the prediction layer on top of each token \texttt{<y>} to predict the function value given the tokens preceding it. Each prediction consists of a mean and a standard deviation value which will be used for the selection of candidates during optimization. Figure~\ref{fig:architecture} illustrates the architecture of \name{}.

It is worth noting that the combination of natural language and domain-specific embeddings is the main distinction between \name{} and 
previous works such as FPT~\citep{lu2022frozen} which applies pretrained LLMs to \textit{sequence classification} tasks in non-language modalities. FPT also learns new embedding layers for the new domain, but relies entirely on the pretrained self-attention layers to model these embeddings without any language instructions.
This distinction stems from the different nature of the tasks we aim to tackle. In sequence classification, the model produces a single prediction for the entire sequence, thus having a good representation of the sequence via self-attention is sufficient. For in-context learning, however, the model must associate each input $x$ with its value $y$ to infer the underlying function $f$ and make predictions for unknown $y$. A language instruction that specifies where $x$ is and where $y$ is helps the model identify this association and improve its in-context reasoning.
Our ablation study in~\ref{sec:language} confirms this utility of retaining language tokens.

\subsection{Semi-synthetic Training}
Our goal is to train \name{} on the unlabeled data $\mathcal{D}_{\text{u}}$ with an unsupervised objective to facilitate efficient generalization to an arbitrary objective function $f$ in the same domain $\mathcal{X}$ after training. Our key insight is that if we train the model to perform the estimation in~\eqref{eq:estimation} for a wide range of functions, it should adapt to any objective function post-training. While the true function values are unknown before optimization, we can use the unlabeled data $x's$ to generate training data from \textit{other functions}. Assume we have access to a family of functions $\Tilde{\mathcal{F}}$ that operate on the same input domain $\mathcal{X}$. For each function $\Tilde{f}$ drawn from $\Tilde{\mathcal{F}}$, we sample a set of function evaluations $\{(x_i,y_i)\}_{i=1}^n$ and train the model to autoregressively predict $y$ given the input $x$ and preceding $(x,y)$ pairs:
\begin{equation}
    \begin{aligned}
        \mathcal{L}(\theta) = \mathbb{E} \left [\sum_{i=1}^n \log p_\theta(y_i \mid x_i, x_{<i}, y_{<i}) \right],
    \end{aligned}
\end{equation}
in which the expectation is with respect to $\Tilde{f} \sim \Tilde{\mathcal{F}}$, $x_{1:n} \sim \mathcal{D}_{\text{u}}$, and $y_{1:n} = \Tilde{f}(x_{1:n})$. After training, the estimation in~\eqref{eq:estimation} can be done purely via in-context prompting, where we condition the model on past observations to make predictions for new data points.

Ideally, the function family $\Tilde{\mathcal{F}}$ should be close to the downstream objective $f$, but also be diverse enough to encourage broad generalization across functions. To achieve this, we propose to train \name{} on a mix of \textit{intrinsic} and \textit{synthetic} functions, which we term \textit{semi-synthetic} training. Intrinsic functions are functions that map each input molecule $x$ to an inherent property of $x$. For example, molecular weight, the number of rings, or heavy atom count are intrinsic properties of the molecule that are known from domain knowledge or can be easily computed using standard tools. These intrinsic properties are closely related to many downstream objective functions. For example, the biological activity of a drug molecule, such as its ability to inhibit a particular enzyme, is often closely related to the molecule's shape or conformation. Therefore, training \name{} from these functions encourages the model to learn useful representations of the input $x$ and obtain good prior knowledge about the optimization domain.

However, it is important to note that we are ultimately interested in optimizing other functions outside of the intrinsic function set. Training the model only on a limited set of intrinsic functions may result in overfitting and poor generalization to unseen functions. To diversify the training data, we additionally train the model on synthetically generated functions. A synthetic function family should be easy to sample from and be capable of producing diverse functions. Many such families exist, including Gaussian Processes (GPs), randomly constructed Gaussian Mixture Models, or randomly initialized neural networks. We choose to generate synthetic functions from Gaussian Processes with a Tanimoto kernel due to its simplicity and efficiency. Tanimoto kernel, also known as the Jaccard coefficient, measures the similarity between two vectors of binary values, a representation that is widely used for many scientific domains such as chemistry, drug discovery, or bioinformatics. Specifically, each synthetic function $\Tilde{f}$ is sampled as follows,
\begin{equation}
    \Tilde{f} \sim \mathcal{G}\mathcal{P}(0, \mathcal{K}), \hspace{0.2cm} \mathcal{K}(x, x') = \frac{x \cdot x'}{||x||^2 + ||x'||^2 - x \cdot x'}, \label{eq:synthetic_sampling}
\end{equation}
where $\mathcal{K}(x, x')$ is the Tanimoto kernel that measures the similarity between $x$ and $x'$.

The final family of functions $\Tilde{\mathcal{F}}$ used to train \name{} is a mixture of intrinsic and synthetic functions with a certain ratio. This design choice is critical to the model's performance. Intuitively, training on both types of functions ensures proximity to the downstream objectives and good coverage of the function space for efficient generalization.
The use of intrinsic functions is also the main difference between our work and ExPT~\citep{nguyen2023expt}, a recent method that studies pure synthetic pretraining for optimization. We hypothesize that while synthetic data is sufficient for ExPT on a few simple tasks, for a more complex domain such as molecular optimization, synthetic training provides too little relevant signal for the model to generalize to downstream objectives. We empirically show the importance of both intrinsic and synthetic functions in the ablation study in section~\ref{sec:synthetic_ratio}.

\subsection{\name{} for Black-box Optimization}
After training, a single \name{} model can be used for optimizing various objective functions within the domain $\mathcal{X}$. Optimization involves an iterative process. At each iteration $t$, we generate a set of candidates $\{x_i\}_{i=1}^C$ using standard crossover and mutation operations for which the model predicts the mean $\mu_i$ and standard deviation $\sigma_i$ conditioned on prior observations $\mathcal{D}_{\text{obs}}$, a dataset of $(x,y)$ pairs collected until $t$. An acquisition function $\alpha$ then calculates a utility score based on $\mu_i$ and $\sigma_i$ for each candidate, balancing between exploration (favoring high $\sigma$) and exploitation (favoring high $\mu$). 
The top $k$ candidates determined by their utility scores are evaluated using the objective function $f$. These $k$ candidates and their corresponding evaluations are incorporated into the dataset $\mathcal{D}_{\text{obs}}$, and the cycle repeats. This process terminates once we exhaust the evaluation budget of $B$. 
Algorithm~\ref{alg:optimization} summarizes the optimization process and Appendix~\ref{app:bbo_hps} outlines the optimization hyperparameters.
\section{Experiments} \label{sec:experiments}

We evaluate \name{} on molecular optimization, where the goal is to design new molecules with desired properties such as high chemical stability, low toxicity, or selective inhibition against a target disease. This problem plays a pivotal role in advancing drug and material discovery. 

\subsection{PMO Benchmark}

\textbf{Benchmark } We evaluate \name{} on Practical Molecular Optimization (PMO)~\citep{gao2022sample}, a standard benchmark for molecular optimization with a focus on sample efficiency. We experiment on $23$ optimization objectives provided by PMO, including QED~\cite{bickerton2012quantifying}, DRD2~\citep{olivecrona2017molecular}, GSK3$\beta$, JNK3~\citep{li2018multi}, and $19$ objective functions from Guacamol~\citep{brown2019guacamol}. QED assesses a molecule's drug-likeness by identifying certain "red flags". DRD2 is a machine learning model trained on experimental data to predict bioactivities for specific target diseases. Guacamol objectives emulate drug discovery goals through a multi-property objective (MPO) approach, considering factors like target molecule similarity, molecular weights, and CLogP. All objective values range from $0$ to $1$, with $1$ indicating the best outcome. We consider two evaluation settings -- the original PMO with a budget of $10000$ oracle calls, and a budget-efficient setting with $1000$ oracle calls, which we refer to as PMO-1K. We believe 1000 is a more reasonable budget while still allowing optimization methods to achieve meaningful performances. To ensure a fair comparison in PMO-1K, we performed extensive hyperparameter tuning for each baseline on the first $5$ tasks, and used the optimal hyperparameters for the remaining tasks. Appendix~\ref{app:hp_tuning} details hyperparameter search for the baselines.

\textbf{Baselines } We compare \name{} against $6$ leading methods in PMO, namely Genetic GFN~\citep{kim2024genetic}, REINVENT~\citep{olivecrona2017molecular}, Augmented Memory~\citep{guo2023augmented}, Graph GA~\citep{jensen2019graph}, GP BO~\citep{tripp2021fresh}, and MOLLEO~\citep{wang2024efficient}. 
Genetic GFN employs a GFlowNets~\citep{bengio2023gflownet} model trained to sample molecules proportional to their rewards.
REINVENT is a reinforcement learning method that finetunes a pretrained RNN for generating SMILES strings, and Augmented Memory combines REINVENT with data augmentation and experience replay. Graph GA, inspired by evolutionary processes, utilizes crossover and mutation operations to explore the molecule space. GP BO is a Bayesian optimization method that augments Graph GA with a Gaussian Processes surrogate model and UCB acquisition function to guide candidate selection.
MOLLEO is an LLM-based method that prompts a chemistry-aware LLM such as BioT5~\citep{pei2023biot5} to perform mutation and crossover operations in an evolutionary algorithm.
Among the baselines, GP BO is the most similar to \name{}, where the only difference is we use an LLM for surrogate modeling instead of a GP.

\textbf{\name{} training } We use ZINC 250K as the unlabeled dataset $\mathcal{D}_{\text{u}}$. ZINC 250K contains around $250000$ molecules sampled from the full ZINC database~\citep{sterling2015zinc} with moderate size and high pharmaceutical relevance and popularity. We adopt $2$-radius $2048$ bit Morgan molecular fingerprints as the input feature of the molecule. To generate training data, we use $47$ intrinsic properties of the molecule as the intrinsic functions, which we present in detail in Appendix~\ref{sec:intrinsic_props}. We train \name{} for $20000$ iterations with a batch size of $4$, where each data point is a sequence of $(x,y)$ pairs sampled from an intrinsic or synthetic function. The ratio of synthetic data is $0.1$. We use Llama-2-7b~\citep{touvron2023llama2} as the base LLM, and use LoRA~\citep{hu2021lora} for parameter-efficient finetuning. We use the \texttt{Llama-2-7b-chat} checkpoint for the 1000-budget setting, and the \texttt{Llama-2-7B-32K-Instruct} checkpoint with the Liger Kernel~\citep{hsu2024ligerkernelefficienttriton} for training and inference with long context for the 10000-budget setting.

\textbf{Evaluation details } We report the area under the curve (AUC) of the top-$10$ average objective value against the number of function calls as the performance metric. AUC metric favors methods that obtain high values with a smaller number of function calls, thus evaluating both optimization capability and sample efficiency. We min-max scale the AUC values to $[0, 1]$. We aggregate the performance for each method across $5$ seeds for better reproducibility as suggested by PMO.

\begin{table*}[t!]
    \centering
    \caption{The performance of \name{} and the baselines on $23$ optimization tasks in PMO-1K. A higher score is better. We report the mean and stddev of scores averaged over $5$ random seeds. We use \textcolor{blue}{\textbf{blue}} and \textcolor{violet}{\textbf{violet}} to denote the best and second-best method for each task. 
    }
    \label{tab:main_tab}
    \resizebox{1.0\linewidth}{!}{
    \begin{tabular}{cccccccc}
        \toprule
        Task & GP BO & Graph GA & REINVENT & \name{} & Genetic GFN & Augmented Memory & MOLLEO \\
        \midrule
        \texttt{albuterol\_similarity} & $0.636 \pm 0.106$ & $0.583 \pm 0.065$ & $0.496 \pm 0.020$ & $0.656 \pm 0.125$ & $\violet{0.664 \pm 0.054}$ & $0.557 \pm 0.048$ & $\blue{0.886 \pm 0.023}$ \\
        \texttt{amlodipine\_mpo} & $0.519 \pm 0.014$ & $0.501 \pm 0.016$ & $0.472 \pm 0.008$ & $\violet{0.541 \pm 0.026}$ & $\violet{0.534 \pm 0.019}$ & $0.489 \pm 0.009$ & $\blue{0.637 \pm 0.023}$ \\
        \texttt{celecoxib\_rediscovery} & $0.411 \pm 0.046$ & $0.424 \pm 0.049$ & $0.370 \pm 0.029$ & $\violet{0.447 \pm 0.073}$ & $\blue{0.447 \pm 0.028}$ & $0.385 \pm 0.027$ & $0.402 \pm 0.003$ \\
        \texttt{deco\_hop} & $0.593 \pm 0.013$ & $0.581 \pm 0.006$ & $0.572 \pm 0.006$ & $\violet{0.596 \pm 0.010}$ & $\blue{0.604 \pm 0.017}$ & $0.579 \pm 0.010$ & $0.588 \pm 0.007$ \\
        \texttt{drd2} & $0.857 \pm 0.080$ & $0.833 \pm 0.065$ & $0.775 \pm 0.086$ & $\violet{0.859 \pm 0.066}$ & $0.809 \pm 0.045$ & $0.795 \pm 0.024$ & $\blue{0.910 \pm 0.017}$ \\
        \texttt{fexofenadine\_mpo} & $\blue{0.707 \pm 0.021}$ & $0.666 \pm 0.009$ & $0.650 \pm 0.007$ & $\violet{0.700 \pm 0.023}$ & $0.682 \pm 0.021$ & $0.679 \pm 0.021$ & $0.674 \pm 0.002$ \\
        \texttt{gsk3b} & $0.611 \pm 0.059$ & $0.523 \pm 0.047$ & $0.589 \pm 0.063$ & $\violet{0.617 \pm 0.063}$ & $\blue{0.637 \pm 0.018}$ & $0.539 \pm 0.097$ & $0.397 \pm 0.013$ \\
        \texttt{isomers\_c7h8n2o2} & $0.545 \pm 0.158$ & $0.735 \pm 0.112$ & $0.725 \pm 0.064$ & $\blue{0.779 \pm 0.099}$ & $\violet{0.738 \pm 0.039}$ & $0.661 \pm 0.039$ & $0.737 \pm 0.043$ \\
        \texttt{isomers\_c9h10n2o2pf2cl} & $0.599 \pm 0.059$ & $0.630 \pm 0.086$ & $0.630 \pm 0.032$ & $\blue{0.672 \pm 0.075}$ & $\violet{0.656 \pm 0.075}$ & $0.596 \pm 0.066$ & $0.635 \pm 0.017$ \\
        \texttt{jnk3} & $\violet{0.346 \pm 0.067}$ & $0.301 \pm 0.071$ & $0.315 \pm 0.042$ & $0.336 \pm 0.051$ & $\blue{0.409 \pm 0.165}$ & $0.294 \pm 0.110$ & $0.186 \pm 0.076$ \\
        \texttt{median1} & $0.213 \pm 0.020$ & $0.208 \pm 0.015$ & $0.205 \pm 0.012$ & $0.217 \pm 0.019$ & $\violet{0.219 \pm 0.008}$ & $0.219 \pm 0.014$ & $\violet{0.236 \pm 0.021}$ \\
        \texttt{median2} & $\violet{0.203 \pm 0.009}$ & $0.181 \pm 0.009$ & $0.188 \pm 0.010$ & $0.193 \pm 0.009$ & $\blue{0.204 \pm 0.011}$ & $0.184 \pm 0.010$ & $0.191 \pm 0.009$ \\
        \texttt{mestranol\_similarity} & $\blue{0.427 \pm 0.025}$ & $0.362 \pm 0.017$ & $0.379 \pm 0.026$ & $\violet{0.423 \pm 0.016}$ & $0.414 \pm 0.022$ & $0.393 \pm 0.021$ & $0.399 \pm 0.020$ \\
        \texttt{osimertinib\_mpo} & $\violet{0.766 \pm 0.006}$ & $0.751 \pm 0.005$ & $0.737 \pm 0.007$ & $0.759 \pm 0.008$ & $0.763 \pm 0.008$ & $0.761 \pm 0.006$ & $\blue{0.779 \pm 0.006}$ \\
        \texttt{perindopril\_mpo} & $0.458 \pm 0.019$ & $0.435 \pm 0.016$ & $0.404 \pm 0.009$ & $\violet{0.473 \pm 0.009}$ & $0.462 \pm 0.033$ & $0.422 \pm 0.013$ & $\blue{0.655 \pm 0.054}$ \\
        \texttt{qed} & $0.912 \pm 0.010$ & $0.914 \pm 0.007$ & $0.921 \pm 0.002$ & $\violet{0.925 \pm 0.005}$ & $\blue{0.928 \pm 0.002}$ & $0.923 \pm 0.002$ & $0.919 \pm 0.006$ \\
        \texttt{ranolazine\_mpo} & $\blue{0.701 \pm 0.023}$ & $0.620 \pm 0.014$ & $0.574 \pm 0.044$ & $\violet{0.687 \pm 0.029}$ & $0.623 \pm 0.022$ & $0.614 \pm 0.033$ & $0.640 \pm 0.000$ \\
        \texttt{scaffold\_hop} & $0.478 \pm 0.009$ & $0.461 \pm 0.008$ & $0.447 \pm 0.010$ & $\violet{0.480 \pm 0.008}$ & $\blue{0.485 \pm 0.015}$ & $0.460 \pm 0.010$ & $0.473 \pm 0.000$ \\
        \texttt{sitagliptin\_mpo} & $0.232 \pm 0.083$ & $0.229 \pm 0.053$ & $\violet{0.261 \pm 0.026}$ & $\blue{0.315 \pm 0.097}$ & $0.227 \pm 0.041$ & $0.245 \pm 0.030$ & $0.193 \pm 0.073$ \\
        \texttt{thiothixene\_rediscovery} & $0.351 \pm 0.039$ & $0.322 \pm 0.023$ & $0.311 \pm 0.021$ & $0.343 \pm 0.035$ & $\violet{0.377 \pm 0.015}$ & $0.336 \pm 0.033$ & $\blue{0.416 \pm 0.075}$ \\
        \texttt{troglitazone\_rediscovery} & $\blue{0.313 \pm 0.018}$ & $0.267 \pm 0.015$ & $0.246 \pm 0.009$ & $0.292 \pm 0.028$ & $0.277 \pm 0.015$ & $0.262 \pm 0.012$ & $\violet{0.302 \pm 0.022}$ \\
        \texttt{valsartan\_smarts} & $0.000 \pm 0.000$ & $0.000 \pm 0.000$ & $0.000 \pm 0.000$ & $0.000 \pm 0.000$ & $0.000 \pm 0.000$ & $0.000 \pm 0.000$ & $0.000 \pm 0.000$ \\
        \texttt{zaleplon\_mpo} & $0.392 \pm 0.034$ & $0.374 \pm 0.024$ & $\violet{0.406 \pm 0.017}$ & $0.404 \pm 0.022$ & $0.400 \pm 0.014$ & $\blue{0.415 \pm 0.013}$ & $0.392 \pm 0.003$ \\
        \midrule
        Sum of scores ($\uparrow$) & $11.27$ & $10.90$ & $10.68$ & $\blue{11.71}$ & $11.56$ & $10.81$ & $\violet{11.65}$ \\
        \bottomrule
    \end{tabular}
    }
\end{table*}

\begin{table*}[t!]
    \centering
    \caption{The performance of \name{} and the baselines on $23$ optimization tasks in PMO. A higher score is better. We report the mean and stddev of scores averaged over $5$ random seeds. We use \textcolor{blue}{\textbf{blue}} and \textcolor{violet}{\textbf{violet}} to denote the best and second-best method for each task.}
    \label{tab:main_tab_10k}
    \resizebox{1.0\linewidth}{!}{
\begin{tabular}{cccccccc}
\toprule
Task & GP BO & Graph GA & REINVENT & LICO & Genetic GFN & Augmented Memory & MOLLEO \\
\midrule
\texttt{albuterol\_similarity} & $0.898 \pm 0.014$ & $0.838 \pm 0.016$ & $0.882 \pm 0.006$ & $0.885 \pm 0.019$ & $\blue{0.941 \pm 0.021}$ & $0.913 \pm 0.009$ & $\violet{0.936 \pm 0.016}$ \\
\texttt{amlodipine\_mpo} & $0.583 \pm 0.044$ & $0.661 \pm 0.020$ & $0.635 \pm 0.035$ & $0.679 \pm 0.027$ & $\violet{0.709 \pm 0.027}$ & $0.691 \pm 0.047$ & $\blue{0.801 \pm 0.028}$ \\
\texttt{celecoxib\_rediscovery} & $0.723 \pm 0.053$ & $0.630 \pm 0.097$ & $0.713 \pm 0.067$ & $0.664 \pm 0.122$ & $\violet{0.784 \pm 0.032}$ & $\blue{0.796 \pm 0.008}$ & $0.459 \pm 0.080$ \\
\texttt{deco\_hop} & $0.629 \pm 0.018$ & $0.619 \pm 0.004$ & $\blue{0.666 \pm 0.044}$ & $0.619 \pm 0.015$ & $0.653 \pm 0.028$ & $\violet{0.658 \pm 0.024}$ & $0.648 \pm 0.099$ \\
\texttt{drd2} & $0.923 \pm 0.017$ & $\blue{0.964 \pm 0.012}$ & $0.945 \pm 0.007$ & $0.928 \pm 0.018$ & $\violet{0.963 \pm 0.006}$ & $\violet{0.963 \pm 0.006}$ & $0.962 \pm 0.013$ \\
\texttt{fexofenadine\_mpo} & $0.722 \pm 0.005$ & $0.760 \pm 0.011$ & $0.784 \pm 0.006$ & $0.772 \pm 0.023$ & $\blue{0.793 \pm 0.009}$ & $\blue{0.859 \pm 0.009}$ & $0.776 \pm 0.019$ \\
\texttt{gsk3b} & $0.851 \pm 0.041$ & $0.788 \pm 0.070$ & $0.865 \pm 0.043$ & $\violet{0.876 \pm 0.045}$ & $0.861 \pm 0.022$ & $\blue{0.881 \pm 0.021}$ & $0.865 \pm 0.037$ \\
\texttt{isomers\_c7h8n2o2} & $0.680 \pm 0.117$ & $0.862 \pm 0.065$ & $0.852 \pm 0.036$ & $\violet{0.939 \pm 0.022}$ & $\blue{0.955 \pm 0.007}$ & $0.853 \pm 0.087$ & $0.915 \pm 0.036$ \\
\texttt{isomers\_c9h10n2o2pf2cl} & $0.469 \pm 0.180$ & $0.719 \pm 0.047$ & $0.642 \pm 0.054$ & $\violet{0.819 \pm 0.039}$ & $\blue{0.876 \pm 0.018}$ & $0.736 \pm 0.051$ & $0.708 \pm 0.093$ \\
\texttt{jnk3} & $0.564 \pm 0.155$ & $0.553 \pm 0.136$ & $\blue{0.783 \pm 0.023}$ & $0.731 \pm 0.037$ & $\violet{0.759 \pm 0.063}$ & $0.739 \pm 0.110$ & $0.715 \pm 0.026$ \\
\texttt{median1} & $0.301 \pm 0.014$ & $0.294 \pm 0.021$ & $\blue{0.356 \pm 0.009}$ & $0.291 \pm 0.016$ & $\violet{0.355 \pm 0.009}$ & $0.326 \pm 0.013$ & $0.302 \pm 0.031$ \\
\texttt{median2} & $\blue{0.297 \pm 0.009}$ & $0.273 \pm 0.009$ & $0.276 \pm 0.008$ & $0.280 \pm 0.019$ & $0.289 \pm 0.007$ & $\violet{0.291 \pm 0.008}$ & $0.206 \pm 0.015$ \\
\texttt{mestranol\_similarity} & $0.627 \pm 0.089$ & $0.579 \pm 0.022$ & $0.618 \pm 0.048$ & $0.614 \pm 0.064$ & $0.697 \pm 0.035$ & $\violet{0.750 \pm 0.049}$ & $\blue{0.759 \pm 0.102}$ \\
\texttt{osimertinib\_mpo} & $0.787 \pm 0.006$ & $0.831 \pm 0.005$ & $0.837 \pm 0.009$ & $0.820 \pm 0.012$ & $\violet{0.846 \pm 0.008}$ & $\blue{0.855 \pm 0.004}$ & $0.819 \pm 0.020$ \\
\texttt{perindopril\_mpo} & $0.493 \pm 0.011$ & $0.538 \pm 0.009$ & $0.537 \pm 0.016$ & $0.557 \pm 0.028$ & $0.595 \pm 0.011$ & $\violet{0.613 \pm 0.015}$ & $\blue{0.723 \pm 0.018}$ \\
\texttt{qed} & $0.937 \pm 0.000$ & $0.940 \pm 0.000$ & $\violet{0.941 \pm 0.000}$ & $0.936 \pm 0.001$ & $0.937 \pm 0.000$ & $\blue{0.942 \pm 0.000}$ & $0.933 \pm 0.003$ \\
\texttt{ranolazine\_mpo} & $0.735 \pm 0.013$ & $0.728 \pm 0.012$ & $0.760 \pm 0.009$ & $0.774 \pm 0.008$ & $\blue{0.810 \pm 0.011}$ & $\violet{0.801 \pm 0.006}$ & $0.731 \pm 0.023$ \\
\texttt{scaffold\_hop} & $0.548 \pm 0.019$ & $0.517 \pm 0.007$ & $0.560 \pm 0.019$ & $0.547 \pm 0.026$ & $\blue{0.585 \pm 0.041}$ & $\violet{0.567 \pm 0.008}$ & $0.516 \pm 0.022$ \\
\texttt{sitagliptin\_mpo} & $0.186 \pm 0.055$ & $0.433 \pm 0.075$ & $0.021 \pm 0.003$ & $\violet{0.567 \pm 0.034}$ & $\blue{0.577 \pm 0.036}$ & $0.284 \pm 0.050$ & $0.496 \pm 0.020$ \\
\texttt{thiothixene\_rediscovery} & $0.559 \pm 0.027$ & $0.479 \pm 0.025$ & $0.534 \pm 0.013$ & $0.514 \pm 0.037$ & $\violet{0.599 \pm 0.073}$ & $0.550 \pm 0.041$ & $\blue{0.658 \pm 0.024}$ \\
\texttt{troglitazone\_rediscovery} & $0.410 \pm 0.015$ & $0.390 \pm 0.016$ & $0.441 \pm 0.032$ & $0.380 \pm 0.026$ & $\violet{0.455 \pm 0.016}$ & $\blue{0.540 \pm 0.048}$ & $0.352 \pm 0.040$ \\
\texttt{valsartan\_smarts} & $0.000 \pm 0.000$ & $0.000 \pm 0.000$ & $0.178 \pm 0.358$ & $0.000 \pm 0.000$ & $\blue{0.092 \pm 0.242}$ & $0.000 \pm 0.000$ & $0.000 \pm 0.000$ \\
\texttt{zaleplon\_mpo} & $0.221 \pm 0.072$ & $0.346 \pm 0.032$ & $0.358 \pm 0.062$ & $\violet{0.515 \pm 0.017}$ & $\blue{0.545 \pm 0.023}$ & $0.394 \pm 0.026$ & $0.402 \pm 0.019$ \\
\midrule
Sum of scores ($\uparrow$) & $13.156$ & $13.751$ & $14.196$ & $14.708$ & $\blue{15.678}$ & $\violet{15.002}$ & $14.682$ \\
\bottomrule
\end{tabular}
}
\end{table*}

\textbf{Results } Table~\ref{tab:main_tab} summarizes the performance of the $7$ considered methods across $23$ optimization tasks in PMO-1K. Overall, \name{} is the leading method in this benchmark, achieving the highest aggregated score. Specifically, \name{} achieves the best or second-best performance in $14/23$ tasks. MOLLEO performs competitively with LICO on this benchmark, with a sum score of $11.65$. However, we note that MOLLEO has significant advantages over LICO and other methods. MOLLEO used BioT5 to generate valid molecules, which has been finetuned extensively on molecules, protein, and molecule-related text data, while LICO leveraged a general LLM like Llama. Moreover, MOLLEO prompts the LLM with a detailed textual description of the task such as \textit{"Your job is to generate a SELFIES molecule that looks more like a drug"}, which possibly has data contamination issues, since the finetuning data may have included similar tasks. On the other hand, we use LICO as a black-box surrogate model, where the model makes predictions based purely on in-context learning of the mapping between molecule fingerprints and their corresponding scores.

On the original PMO setting, Table~\ref{tab:main_tab_10k} shows the competitive performance of \name{} with the two state-of-the-art methods Genetic GFN and Augmented Memory. It is important to note that other methods have significant advantages over \name{}, since both Genetic GFN and Augmented Memory update their models from real data during optimization, whereas \name{} performs in-context learning without being explicitly trained on data from downstream objectives. This impressive result shows the effectiveness of semi-synthetic training in enabling generalization to a broad range of functions via in-context prompting. 

\begin{figure*}[t]
    \centering
    \includegraphics[width=0.85\linewidth]{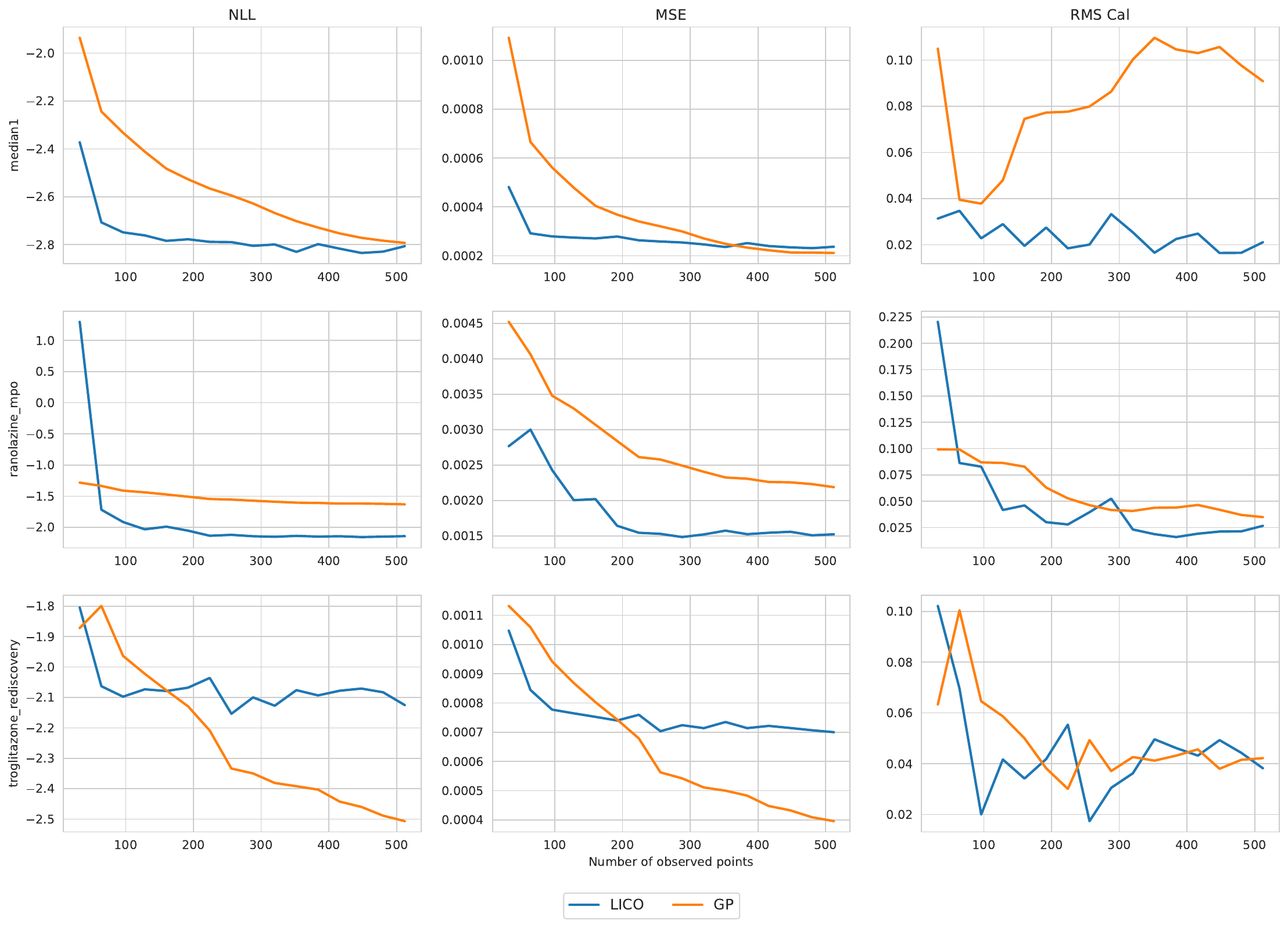}
    \caption{The predictive performance of \name{} and GP on $3$ objective functions in PMO with different metrics and varying numbers of observations.}
    \label{fig:prediction_compare}
\end{figure*}

The most closely related method to \name{} is GP BO, where the only difference between the two is the surrogate model. This indicates the superiority of \name{} compared to GP, a popular surrogate model for black-box optimization. To verify this, we compare the predictive performance of \name{} and GP on several objective functions. We do this by first labeling the ZINC unlabeled dataset with the objective functions and randomly choosing a subset of the labeled data points for evaluation. For each task, we vary the number of examples given to each method from $32$ to $512$, and evaluate their performance on $128$ held-out data points. We use negative log-likelihood, mean squared error, and root mean squared calibration error as the evaluation metrics. 
Figure~\ref{fig:prediction_compare} compares the predictive performance of \name{} and GP in $3$ objective functions, \texttt{median1}, \texttt{ranolazine\_mpo}, and \texttt{troglitazone\_rediscovery}. The figure shows that the optimization performance of the method closely aligns with the predictive performance of the surrogate model. In \texttt{median1} and \texttt{ranolazine\_mpo} where \name{} outperforms GP in terms of optimization score, the model also achieves lower negative log-likelihood, mean squared error, and calibration error. Similarly, \name{} has worse predictive performance in \texttt{troglitazone\_rediscovery} where it underperforms GP. This verifies our hypothesis and proves the effectiveness of \name{} for surrogate modeling.

\begin{table*}[t!]
    \centering
    \caption{Performance of \name{} on $5$ tasks with different language instructions.}
    \label{tab:language}
    \resizebox{0.99\linewidth}{!}{
    \begin{tabular}{ccccccc}
        \toprule
        Task & \texttt{albuterol\_similarity} & \texttt{amlodipine\_mpo} & \texttt{celecoxib\_rediscovery} & \texttt{deco\_hop} & \texttt{drd2} & Sum ($\uparrow$) \\
        \midrule

        \name{} w/o Language & $0.615 \pm 0.104$ & $0.491 \pm 0.018$ & $0.396 \pm 0.051$ & $0.585 \pm 0.010$ & $0.840 \pm 0.063$ & $2.927$ \\
        
        \name{} w/o Task prompt & $0.641 \pm 0.107$ & $0.523 \pm 0.018$ & $\mathbf{0.457 \pm 0.041}$ & $0.595 \pm 0.006$ & $0.844 \pm 0.105$ & $3.060$ \\
        
        \name{} & $\mathbf{0.656 \pm 0.125}$ & $\mathbf{0.541 \pm 0.026}$ & $0.447 \pm 0.073$ & $\mathbf{0.596 \pm 0.010}$ & $\mathbf{0.859 \pm 0.066}$ & $\mathbf{3.099}$ \\
        
        \bottomrule
    \end{tabular}
    }
\end{table*}

\begin{table*}[t!]
    \centering
    \caption{Performance of \name{} on $5$ tasks with different ratios of synthetic data.}
    \label{tab:ablation_ratio}
    \resizebox{0.99\linewidth}{!}{
    \begin{tabular}{ccccccc}
        \toprule
        Task & \texttt{albuterol\_similarity} & \texttt{amlodipine\_mpo} & \texttt{celecoxib\_rediscovery} & \texttt{deco\_hop} & \texttt{drd2} & Sum ($\uparrow$) \\
        \midrule

        \name{} Intrinsic & $0.598 \pm 0.115$ & $0.524 \pm 0.029$ & $0.412 \pm 0.042$ & $0.585 \pm 0.005$ & $0.891 \pm 0.032$ & $3.010$ \\
        
        \name{} 0.1 Synthetic & $0.656 \pm 0.125$ & $\mathbf{0.541 \pm 0.026}$ & $\mathbf{0.447 \pm 0.073}$ & $\mathbf{0.596 \pm 0.010}$ & $0.859 \pm 0.066$ & $\mathbf{3.099}$ \\
        
        \name{} 0.5 Synthetic & $\mathbf{0.663 \pm 0.140}$ & $0.504 \pm 0.016$ & $0.402 \pm 0.016$ & $0.588 \pm 0.006$ & $\mathbf{0.907 \pm 0.020}$ & $3.063$ \\
        
        \name{} Synthetic & $0.547 \pm 0.080$ & $0.498 \pm 0.026$ & $0.404 \pm 0.103$ & $0.585 \pm 0.003$ & $0.902 \pm 0.012$ & $2.936$ \\
        
        \bottomrule
    \end{tabular}
    }
\end{table*}

\subsection{Ablation Analysis}
We perform various ablation studies to understand the importance of different components and design choices in \name{}. For the ablation experiments, we consider the first $5$ tasks in Table~\ref{tab:main_tab} only. We report the aggregated performance of different models using AUC Top-$10$ across $5$ random seeds.

\subsubsection{\name{} without language instruction} \label{sec:language}

First, we examine the importance of language instructions to the performance of \name{}. We compare $3$ variants of \name{}: 1) \name{} without any language instruction, 2) \name{} with special tokens \texttt{<x>} and \texttt{<y>} but without a task prompt, and 3) \name{} with both special tokens and the task prompt. Table~\ref{tab:language} compares the performance of the $3$ variants. \name{} performs the best in $4/5$ tasks, followed by \name{} without the task prompt. \name{} without any language instruction performs the worst, often by a large margin. This result confirms the importance of guiding a pretrained LLM with language instruction when applying the model to in-context reasoning in a completely new domain.

\subsubsection{\name{} with different synthetic ratios} \label{sec:synthetic_ratio}
We investigate the importance of training \name{} on both intrinsic and synthetic data. To do this, we gradually increase the ratio of synthetic functions in the training data from $0$ (intrinsic-only) to $1$ (synthetic-only), and compare the performance of \name{} across different ratios. Table~\ref{tab:ablation_ratio} shows that \name{} with semi-synthetic training performs the best, outperforming both intrinsic-only and synthetic-only data. Training with synthetic data only performs the worst, which is expected when synthetic functions generated by a GP do not include any domain knowledge that is encoded by the intrinsic functions. In other words, synthetic data alone provides too little relevant signal for the model to generalize to unseen downstream objectives. Training with intrinsic functions only, on the other hand, results in quite good performances on most tasks. However, in tasks like \texttt{albuterol\_similarity}, semi-synthetic training outperforms this baseline by a large margin. We hypothesize that the underlying objective in \texttt{albuterol\_similarity} is far from the intrinsic functions used to train \name{}, leading to poor generalization. Finally, training with small ($0.1$) to moderate ($0.5$) ratios of synthetic data achieves similarly good performance.

\begin{figure}[t]
    \centering
    \begin{minipage}{0.54\textwidth}
        \centering
        \captionof{table}{Performance of pretrained vs randomly initialized LLMs.}
        \label{tab:pretrain_scratch}
        \resizebox{\linewidth}{!}{
        \begin{tabular}{ccc}
            \toprule
            Task           & Pretrained LLM        & Scratch LLM \\ \midrule
            \texttt{albuterol\_similarity}    & $\mathbf{0.656 \pm 0.125}$ & $0.575 \pm 0.064$ \\
            \texttt{amlodipine\_mpo}          & $\mathbf{0.541 \pm 0.026}$ & $0.503 \pm 0.029$ \\
            \texttt{celecoxib\_rediscovery}   & $\mathbf{0.447 \pm 0.073}$ & $0.410 \pm 0.034$ \\
            \texttt{deco\_hop}                & $\mathbf{0.596 \pm 0.010}$ & $0.583 \pm 0.005$ \\
            \texttt{drd2}            & $\mathbf{0.859 \pm 0.066}$ & $0.827 \pm 0.085$ \\ \midrule
            Sum              & $\mathbf{3.099}$            & $2.898$           \\ \bottomrule
        \end{tabular}
        }
    \end{minipage}
    \hfill
    \begin{minipage}{0.42\textwidth}
        \centering
        \includegraphics[width=\linewidth]{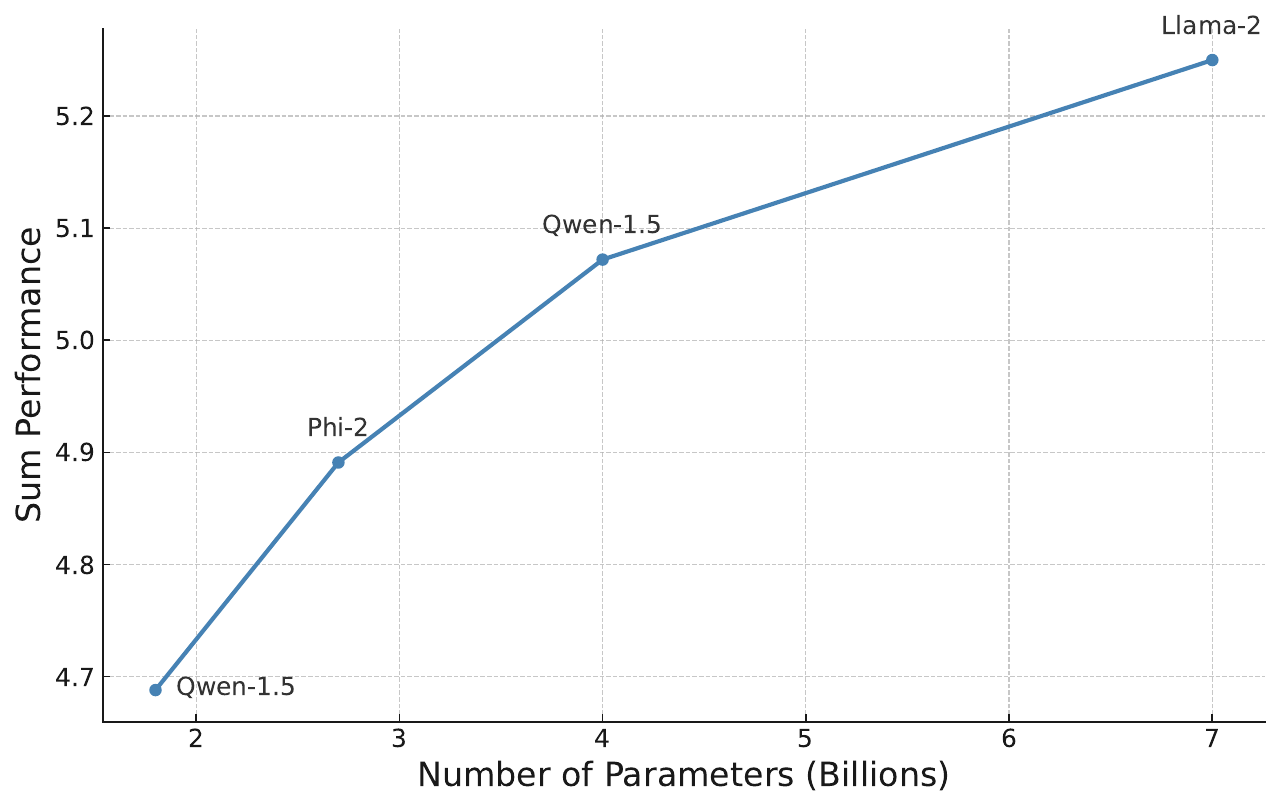}
        \caption{\name{} with different LLM sizes.}
        \label{fig:scaling}
    \end{minipage}
\end{figure}

\subsubsection{Randomly initialized vs Pretrained LLMs}
To understand the importance of using a pretrained LLM, we compare \name{} with an autoregressive transformer model of the same size (7B). The transformer architecture is the same as in~\citep{garg2022can}, and we train this model to perform in-context learning on the semi-synthetic data from scratch. Table~\ref{tab:pretrain_scratch} shows the comparison. The scratch model performs much worse than \name{} with a pretrained LLM on all tasks despite sharing the same number of parameters. This highlights the importance of the pattern-matching capabilities that LLMs like Llama-2 acquire via extensive language pretraining.

\subsubsection{\name{} with different LLM sizes}

Previous works have shown the favorable scaling laws of Large Language Models where larger models consistently perform better on downstream tasks~\citep{kaplan2020scaling}. In this section, we investigate the scaling properties of LLMs but in the context of black-box optimization. Specifically, we compare $4$ different base LLMs with different sizes -- Qwen-1.5 1.8B and 4B~\citep{bai2023qwen}, Phi-2 2.7B~\citep{javaheripi2023phi}, and Llama-2 7B~\citep{touvron2023llama2}. We use the same language instructions for all models. We evaluate each model on the first $8$ tasks in Table~\ref{tab:main_tab} and average the results across $5$ random seeds. We report the sum of performance across $8$ tasks.

The comparison in Figure~\ref{fig:scaling} shows that the optimization performance scales consistently with the model size, with Llama-2 7B being the best method. This experiment indicates that larger LLMs not only perform better in language tasks but also obtain stronger pattern-matching capabilities that can be transferred to a completely different domain. Given this scaling, we can further improve the current performance of \name{} by scaling up the base LLM size.

\section{Conclusion and Future Work} \label{sec:conclusion}
We develop \name{}, a new method that leverages pretrained Large Language Models for black-box optimization. \name{} extends existing LLMs to non-language domains with separate embedding and prediction layers. To enable efficient generalization to various optimization tasks, we train \name{} on a diverse set of semi-synthetic functions for few-shot predictions. \name{} achieves state-of-the-art performance on PMO, a challenging molecular optimization benchmark with over 20 objective functions. Ablation analyses highlight the importance of incorporating language instruction to guide in-context learning and semi-synthetic training for better generalization.
One limitation of our method is the assumption of an accessible set of intrinsic functions. While this is true for molecular optimization, it may not apply to other scientific domains. In such cases, a better synthetic data generation process incorporating domain knowledge is needed to aid generalization.
Future directions include evaluating \name{} in other domains to test its applicability and generality, exploring other prompts that better exploit the capabilities of pretrained LLMs, and using LLMs for other aspects of optimization, such as candidate suggestion or exploration.

\clearpage
\bibliography{iclr2025_conference}

\begin{thebibliography}{72}
\providecommand{\natexlab}[1]{#1}
\providecommand{\url}[1]{\texttt{#1}}
\expandafter\ifx\csname urlstyle\endcsname\relax
  \providecommand{\doi}[1]{doi: #1}\else
  \providecommand{\doi}{doi: \begingroup \urlstyle{rm}\Url}\fi

\bibitem[Achiam et~al.(2023)Achiam, Adler, Agarwal, Ahmad, Akkaya, Aleman, Almeida, Altenschmidt, Altman, Anadkat, et~al.]{achiam2023gpt}
Josh Achiam, Steven Adler, Sandhini Agarwal, Lama Ahmad, Ilge Akkaya, Florencia~Leoni Aleman, Diogo Almeida, Janko Altenschmidt, Sam Altman, Shyamal Anadkat, et~al.
\newblock Gpt-4 technical report.
\newblock \emph{arXiv preprint arXiv:2303.08774}, 2023.

\bibitem[Angermueller et~al.(2020)Angermueller, Dohan, Belanger, Deshpande, Murphy, and Colwell]{angermueller2020model}
Christof Angermueller, David Dohan, David Belanger, Ramya Deshpande, Kevin Murphy, and Lucy Colwell.
\newblock Model-based reinforcement learning for biological sequence design.
\newblock 2020.

\bibitem[Bai et~al.(2023)Bai, Bai, Chu, Cui, Dang, Deng, Fan, Ge, Han, Huang, et~al.]{bai2023qwen}
Jinze Bai, Shuai Bai, Yunfei Chu, Zeyu Cui, Kai Dang, Xiaodong Deng, Yang Fan, Wenbin Ge, Yu~Han, Fei Huang, et~al.
\newblock Qwen technical report.
\newblock \emph{arXiv preprint arXiv:2309.16609}, 2023.

\bibitem[Bengio et~al.(2023)Bengio, Lahlou, Deleu, Hu, Tiwari, and Bengio]{bengio2023gflownet}
Yoshua Bengio, Salem Lahlou, Tristan Deleu, Edward~J Hu, Mo~Tiwari, and Emmanuel Bengio.
\newblock Gflownet foundations.
\newblock \emph{The Journal of Machine Learning Research}, 24\penalty0 (1):\penalty0 10006--10060, 2023.

\bibitem[Berkenkamp et~al.(2016)Berkenkamp, Schoellig, and Krause]{berkenkamp2016safe}
Felix Berkenkamp, Angela~P Schoellig, and Andreas Krause.
\newblock Safe controller optimization for quadrotors with gaussian processes.
\newblock In \emph{2016 IEEE international conference on robotics and automation (ICRA)}, pp.\  491--496. IEEE, 2016.

\bibitem[Bickerton et~al.(2012)Bickerton, Paolini, Besnard, Muresan, and Hopkins]{bickerton2012quantifying}
G~Richard Bickerton, Gaia~V Paolini, J{\'e}r{\'e}my Besnard, Sorel Muresan, and Andrew~L Hopkins.
\newblock Quantifying the chemical beauty of drugs.
\newblock \emph{Nature chemistry}, 4\penalty0 (2):\penalty0 90--98, 2012.

\bibitem[Bradley et~al.(2024)Bradley, Fan, Galanos, Zhou, Scott, and Lehman]{bradley2024openelm}
Herbie Bradley, Honglu Fan, Theodoros Galanos, Ryan Zhou, Daniel Scott, and Joel Lehman.
\newblock The openelm library: Leveraging progress in language models for novel evolutionary algorithms.
\newblock \emph{Genetic Programming Theory and Practice XX. Springer}, 2024.

\bibitem[Brookes et~al.(2019)Brookes, Park, and Listgarten]{brookes2019conditioning}
David Brookes, Hahnbeom Park, and Jennifer Listgarten.
\newblock Conditioning by adaptive sampling for robust design.
\newblock In \emph{International conference on machine learning}, pp.\  773--782. PMLR, 2019.

\bibitem[Brown et~al.(2019)Brown, Fiscato, Segler, and Vaucher]{brown2019guacamol}
Nathan Brown, Marco Fiscato, Marwin~HS Segler, and Alain~C Vaucher.
\newblock Guacamol: benchmarking models for de novo molecular design.
\newblock \emph{Journal of chemical information and modeling}, 59\penalty0 (3):\penalty0 1096--1108, 2019.

\bibitem[Brown et~al.(2020)Brown, Mann, Ryder, Subbiah, Kaplan, Dhariwal, Neelakantan, Shyam, Sastry, Askell, Agarwal, Herbert-Voss, Krueger, Henighan, Child, Ramesh, Ziegler, Wu, Winter, Hesse, Chen, Sigler, Litwin, Gray, Chess, Clark, Berner, McCandlish, Radford, Sutskever, and Amodei]{brown2020language}
Tom~B. Brown, Benjamin Mann, Nick Ryder, Melanie Subbiah, Jared Kaplan, Prafulla Dhariwal, Arvind Neelakantan, Pranav Shyam, Girish Sastry, Amanda Askell, Sandhini Agarwal, Ariel Herbert-Voss, Gretchen Krueger, Tom Henighan, Rewon Child, Aditya Ramesh, Daniel~M. Ziegler, Jeffrey Wu, Clemens Winter, Christopher Hesse, Mark Chen, Eric Sigler, Mateusz Litwin, Scott Gray, Benjamin Chess, Jack Clark, Christopher Berner, Sam McCandlish, Alec Radford, Ilya Sutskever, and Dario Amodei.
\newblock Language models are few-shot learners, 2020.

\bibitem[Bubeck et~al.(2023)Bubeck, Chandrasekaran, Eldan, Gehrke, Horvitz, Kamar, Lee, Lee, Li, Lundberg, et~al.]{bubeck2023sparks}
S{\'e}bastien Bubeck, Varun Chandrasekaran, Ronen Eldan, Johannes Gehrke, Eric Horvitz, Ece Kamar, Peter Lee, Yin~Tat Lee, Yuanzhi Li, Scott Lundberg, et~al.
\newblock Sparks of artificial general intelligence: Early experiments with gpt-4.
\newblock \emph{arXiv preprint arXiv:2303.12712}, 2023.

\bibitem[Chen et~al.(2023)Chen, Dohan, and So]{chen2023evoprompting}
Angelica Chen, David~M Dohan, and David~R So.
\newblock Evoprompting: Language models for code-level neural architecture search.
\newblock \emph{arXiv preprint arXiv:2302.14838}, 2023.

\bibitem[Chen et~al.(2022)Chen, Song, Lee, Wang, Zhang, Dohan, Kawakami, Kochanski, Doucet, Ranzato, et~al.]{chen2022towards}
Yutian Chen, Xingyou Song, Chansoo Lee, Zi~Wang, Richard Zhang, David Dohan, Kazuya Kawakami, Greg Kochanski, Arnaud Doucet, Marc'aurelio Ranzato, et~al.
\newblock Towards learning universal hyperparameter optimizers with transformers.
\newblock \emph{Advances in Neural Information Processing Systems}, 35:\penalty0 32053--32068, 2022.

\bibitem[Dinh et~al.(2022)Dinh, Zeng, Zhang, Lin, Gira, Rajput, Sohn, Papailiopoulos, and Lee]{dinh2022lift}
Tuan Dinh, Yuchen Zeng, Ruisu Zhang, Ziqian Lin, Michael Gira, Shashank Rajput, Jy-yong Sohn, Dimitris Papailiopoulos, and Kangwook Lee.
\newblock Lift: Language-interfaced fine-tuning for non-language machine learning tasks.
\newblock \emph{Advances in Neural Information Processing Systems}, 35:\penalty0 11763--11784, 2022.

\bibitem[Fang et~al.(2023)Fang, Zhang, Chen, Guo, Fan, and Chen]{fang2023domain}
Yin Fang, Ningyu Zhang, Zhuo Chen, Lingbing Guo, Xiaohui Fan, and Huajun Chen.
\newblock Domain-agnostic molecular generation with self-feedback.
\newblock \emph{arXiv preprint arXiv:2301.11259}, 2023.

\bibitem[Gao et~al.(2022)Gao, Fu, Sun, and Coley]{gao2022sample}
Wenhao Gao, Tianfan Fu, Jimeng Sun, and Connor Coley.
\newblock Sample efficiency matters: a benchmark for practical molecular optimization.
\newblock \emph{Advances in Neural Information Processing Systems}, 35:\penalty0 21342--21357, 2022.

\bibitem[Garg et~al.(2022)Garg, Tsipras, Liang, and Valiant]{garg2022can}
Shivam Garg, Dimitris Tsipras, Percy~S Liang, and Gregory Valiant.
\newblock What can transformers learn in-context? a case study of simple function classes.
\newblock \emph{Advances in Neural Information Processing Systems}, 35:\penalty0 30583--30598, 2022.

\bibitem[Gaulton et~al.(2012)Gaulton, Bellis, Bento, Chambers, Davies, Hersey, Light, McGlinchey, Michalovich, Al-Lazikani, et~al.]{gaulton2012chembl}
Anna Gaulton, Louisa~J Bellis, A~Patricia Bento, Jon Chambers, Mark Davies, Anne Hersey, Yvonne Light, Shaun McGlinchey, David Michalovich, Bissan Al-Lazikani, et~al.
\newblock Chembl: a large-scale bioactivity database for drug discovery.
\newblock \emph{Nucleic acids research}, 40\penalty0 (D1):\penalty0 D1100--D1107, 2012.

\bibitem[Gruver et~al.(2023)Gruver, Finzi, Qiu, and Wilson]{gruver2023large}
Nate Gruver, Marc Finzi, Shikai Qiu, and Andrew~Gordon Wilson.
\newblock Large language models are zero-shot time series forecasters.
\newblock \emph{arXiv preprint arXiv:2310.07820}, 2023.

\bibitem[Guevorguian et~al.(2024)Guevorguian, Bedrosian, Fahradyan, Chilingaryan, Khachatrian, and Aghajanyan]{guevorguian2024small}
Philipp Guevorguian, Menua Bedrosian, Tigran Fahradyan, Gayane Chilingaryan, Hrant Khachatrian, and Armen Aghajanyan.
\newblock Small molecule optimization with large language models.
\newblock \emph{arXiv preprint arXiv:2407.18897}, 2024.

\bibitem[Guo \& Schwaller(2023)Guo and Schwaller]{guo2023augmented}
Jeff Guo and Philippe Schwaller.
\newblock Augmented memory: Capitalizing on experience replay to accelerate de novo molecular design.
\newblock \emph{arXiv preprint arXiv:2305.16160}, 2023.

\bibitem[Hamidieh(2018)]{hamidieh2018data}
Kam Hamidieh.
\newblock A data-driven statistical model for predicting the critical temperature of a superconductor.
\newblock \emph{Computational Materials Science}, 154:\penalty0 346--354, 2018.

\bibitem[Hsu et~al.(2024)Hsu, Dai, Kothapalli, Song, Tang, Zhu, Shimizu, Sahni, Ning, and Chen]{hsu2024ligerkernelefficienttriton}
Pin-Lun Hsu, Yun Dai, Vignesh Kothapalli, Qingquan Song, Shao Tang, Siyu Zhu, Steven Shimizu, Shivam Sahni, Haowen Ning, and Yanning Chen.
\newblock Liger kernel: Efficient triton kernels for llm training.
\newblock \emph{arXiv preprint arXiv:2410.10989}, 2024.
\newblock URL \url{https://arxiv.org/abs/2410.10989}.

\bibitem[Hu et~al.(2021)Hu, Shen, Wallis, Allen-Zhu, Li, Wang, Wang, and Chen]{hu2021lora}
Edward~J Hu, Yelong Shen, Phillip Wallis, Zeyuan Allen-Zhu, Yuanzhi Li, Shean Wang, Lu~Wang, and Weizhu Chen.
\newblock Lora: Low-rank adaptation of large language models.
\newblock \emph{arXiv preprint arXiv:2106.09685}, 2021.

\bibitem[Javaheripi et~al.(2023)Javaheripi, Bubeck, Abdin, Aneja, Bubeck, Mendes, Chen, Del~Giorno, Eldan, Gopi, et~al.]{javaheripi2023phi}
Mojan Javaheripi, S{\'e}bastien Bubeck, Marah Abdin, Jyoti Aneja, Sebastien Bubeck, Caio C{\'e}sar~Teodoro Mendes, Weizhu Chen, Allie Del~Giorno, Ronen Eldan, Sivakanth Gopi, et~al.
\newblock Phi-2: The surprising power of small language models.
\newblock \emph{Microsoft Research Blog}, 2023.

\bibitem[Jensen(2019)]{jensen2019graph}
Jan~H Jensen.
\newblock A graph-based genetic algorithm and generative model/monte carlo tree search for the exploration of chemical space.
\newblock \emph{Chemical science}, 10\penalty0 (12):\penalty0 3567--3572, 2019.

\bibitem[Jiang et~al.(2023)Jiang, Sablayrolles, Mensch, Bamford, Chaplot, Casas, Bressand, Lengyel, Lample, Saulnier, et~al.]{jiang2023mistral}
Albert~Q Jiang, Alexandre Sablayrolles, Arthur Mensch, Chris Bamford, Devendra~Singh Chaplot, Diego de~las Casas, Florian Bressand, Gianna Lengyel, Guillaume Lample, Lucile Saulnier, et~al.
\newblock Mistral 7b.
\newblock \emph{arXiv preprint arXiv:2310.06825}, 2023.

\bibitem[Jiang et~al.(2024)Jiang, Sablayrolles, Roux, Mensch, Savary, Bamford, Chaplot, Casas, Hanna, Bressand, et~al.]{jiang2024mixtral}
Albert~Q Jiang, Alexandre Sablayrolles, Antoine Roux, Arthur Mensch, Blanche Savary, Chris Bamford, Devendra~Singh Chaplot, Diego de~las Casas, Emma~Bou Hanna, Florian Bressand, et~al.
\newblock Mixtral of experts.
\newblock \emph{arXiv preprint arXiv:2401.04088}, 2024.

\bibitem[Kaplan et~al.(2020)Kaplan, McCandlish, Henighan, Brown, Chess, Child, Gray, Radford, Wu, and Amodei]{kaplan2020scaling}
Jared Kaplan, Sam McCandlish, Tom Henighan, Tom~B Brown, Benjamin Chess, Rewon Child, Scott Gray, Alec Radford, Jeffrey Wu, and Dario Amodei.
\newblock Scaling laws for neural language models.
\newblock \emph{arXiv preprint arXiv:2001.08361}, 2020.

\bibitem[Kim et~al.(2024)Kim, Kim, Choi, and Park]{kim2024genetic}
Hyeonah Kim, Minsu Kim, Sanghyeok Choi, and Jinkyoo Park.
\newblock Genetic-guided gflownets: Advancing in practical molecular optimization benchmark.
\newblock \emph{arXiv preprint arXiv:2402.05961}, 2024.

\bibitem[Kojima et~al.(2022)Kojima, Gu, Reid, Matsuo, and Iwasawa]{kojima2022large}
Takeshi Kojima, Shixiang~Shane Gu, Machel Reid, Yutaka Matsuo, and Yusuke Iwasawa.
\newblock Large language models are zero-shot reasoners.
\newblock \emph{Advances in neural information processing systems}, 35:\penalty0 22199--22213, 2022.

\bibitem[Krishnamoorthy et~al.(2023{\natexlab{a}})Krishnamoorthy, Mashkaria, and Grover]{krishnamoorthy2022generative}
Siddarth Krishnamoorthy, Satvik~Mehul Mashkaria, and Aditya Grover.
\newblock Generative pretraining for black-box optimization.
\newblock In \emph{ICML}, 2023{\natexlab{a}}.

\bibitem[Krishnamoorthy et~al.(2023{\natexlab{b}})Krishnamoorthy, Mashkaria, and Grover]{krishnamoorthy2023diffusion}
Siddarth Krishnamoorthy, Satvik~Mehul Mashkaria, and Aditya Grover.
\newblock Diffusion models for black-box optimization.
\newblock In \emph{ICML}, 2023{\natexlab{b}}.

\bibitem[Kristiadi et~al.()Kristiadi, Strieth-Kalthoff, Skreta, Poupart, Aspuru-Guzik, and Pleiss]{kristiadisober}
Agustinus Kristiadi, Felix Strieth-Kalthoff, Marta Skreta, Pascal Poupart, Alan Aspuru-Guzik, and Geoff Pleiss.
\newblock A sober look at llms for material discovery: Are they actually good for bayesian optimization over molecules?
\newblock In \emph{Forty-first International Conference on Machine Learning}.

\bibitem[Lehman et~al.(2023)Lehman, Gordon, Jain, Ndousse, Yeh, and Stanley]{lehman2023evolution}
Joel Lehman, Jonathan Gordon, Shawn Jain, Kamal Ndousse, Cathy Yeh, and Kenneth~O Stanley.
\newblock Evolution through large models.
\newblock In \emph{Handbook of Evolutionary Machine Learning}, pp.\  331--366. Springer, 2023.

\bibitem[Li et~al.(2022)Li, Puig, Paxton, Du, Wang, Fan, Chen, Huang, Aky{\"u}rek, Anandkumar, et~al.]{li2022pre}
Shuang Li, Xavier Puig, Chris Paxton, Yilun Du, Clinton Wang, Linxi Fan, Tao Chen, De-An Huang, Ekin Aky{\"u}rek, Anima Anandkumar, et~al.
\newblock Pre-trained language models for interactive decision-making.
\newblock \emph{Advances in Neural Information Processing Systems}, 35:\penalty0 31199--31212, 2022.

\bibitem[Li et~al.(2018)Li, Zhang, and Liu]{li2018multi}
Yibo Li, Liangren Zhang, and Zhenming Liu.
\newblock Multi-objective de novo drug design with conditional graph generative model.
\newblock \emph{Journal of cheminformatics}, 10:\penalty0 1--24, 2018.

\bibitem[Liao et~al.(2019)Liao, Wang, Yang, Lee, Pister, Levine, and Calandra]{liao2019data}
Thomas Liao, Grant Wang, Brian Yang, Rene Lee, Kristofer Pister, Sergey Levine, and Roberto Calandra.
\newblock Data-efficient learning of morphology and controller for a microrobot.
\newblock In \emph{2019 International Conference on Robotics and Automation (ICRA)}, pp.\  2488--2494. IEEE, 2019.

\bibitem[Liu et~al.(2023{\natexlab{a}})Liu, Lin, Wang, Yao, Tong, Yuan, and Zhang]{liu2023large}
Fei Liu, Xi~Lin, Zhenkun Wang, Shunyu Yao, Xialiang Tong, Mingxuan Yuan, and Qingfu Zhang.
\newblock Large language model for multi-objective evolutionary optimization.
\newblock \emph{arXiv preprint arXiv:2310.12541}, 2023{\natexlab{a}}.

\bibitem[Liu et~al.(2023{\natexlab{b}})Liu, Wang, Yang, Wang, Liu, Guo, and Xiao]{liu2023chatgpt}
Shengchao Liu, Jiongxiao Wang, Yijin Yang, Chengpeng Wang, Ling Liu, Hongyu Guo, and Chaowei Xiao.
\newblock Chatgpt-powered conversational drug editing using retrieval and domain feedback.
\newblock \emph{arXiv preprint arXiv:2305.18090}, 2023{\natexlab{b}}.

\bibitem[Liu et~al.(2024)Liu, Wang, Yang, Wang, Liu, Guo, and Xiao]{liu2024conversational}
Shengchao Liu, Jiongxiao Wang, Yijin Yang, Chengpeng Wang, Ling Liu, Hongyu Guo, and Chaowei Xiao.
\newblock Conversational drug editing using retrieval and domain feedback.
\newblock In \emph{The Twelfth International Conference on Learning Representations}, 2024.

\bibitem[Liu et~al.()Liu, Astorga, Seedat, and van~der Schaar]{liularge}
Tennison Liu, Nicol{\'a}s Astorga, Nabeel Seedat, and Mihaela van~der Schaar.
\newblock Large language models to enhance bayesian optimization.
\newblock In \emph{The Twelfth International Conference on Learning Representations}.

\bibitem[Livne et~al.(2024)Livne, Miftahutdinov, Tutubalina, Kuznetsov, Polykovskiy, Brundyn, Jhunjhunwala, Costa, Aliper, Aspuru-Guzik, et~al.]{livne2024nach0}
Micha Livne, Zulfat Miftahutdinov, Elena Tutubalina, Maksim Kuznetsov, Daniil Polykovskiy, Annika Brundyn, Aastha Jhunjhunwala, Anthony Costa, Alex Aliper, Al{\'a}n Aspuru-Guzik, et~al.
\newblock nach0: Multimodal natural and chemical languages foundation model.
\newblock \emph{Chemical Science}, 15\penalty0 (22):\penalty0 8380--8389, 2024.

\bibitem[Lu et~al.(2022)Lu, Grover, Abbeel, and Mordatch]{lu2022frozen}
Kevin Lu, Aditya Grover, Pieter Abbeel, and Igor Mordatch.
\newblock Frozen pretrained transformers as universal computation engines.
\newblock In \emph{Proceedings of the AAAI Conference on Artificial Intelligence}, volume~36, pp.\  7628--7636, 2022.

\bibitem[Ma et~al.(2023)Ma, Liang, Wang, Huang, Bastani, Jayaraman, Zhu, Fan, and Anandkumar]{ma2023eureka}
Yecheng~Jason Ma, William Liang, Guanzhi Wang, De-An Huang, Osbert Bastani, Dinesh Jayaraman, Yuke Zhu, Linxi Fan, and Anima Anandkumar.
\newblock Eureka: Human-level reward design via coding large language models.
\newblock \emph{arXiv preprint arXiv:2310.12931}, 2023.

\bibitem[Meyerson et~al.(2023)Meyerson, Nelson, Bradley, Moradi, Hoover, and Lehman]{meyerson2023language}
Elliot Meyerson, Mark~J Nelson, Herbie Bradley, Arash Moradi, Amy~K Hoover, and Joel Lehman.
\newblock Language model crossover: Variation through few-shot prompting.
\newblock \emph{arXiv preprint arXiv:2302.12170}, 2023.

\bibitem[Mirchandani et~al.(2023)Mirchandani, Xia, Florence, Ichter, Driess, Arenas, Rao, Sadigh, and Zeng]{mirchandani2023large}
Suvir Mirchandani, Fei Xia, Pete Florence, Brian Ichter, Danny Driess, Montserrat~Gonzalez Arenas, Kanishka Rao, Dorsa Sadigh, and Andy Zeng.
\newblock Large language models as general pattern machines.
\newblock \emph{arXiv preprint arXiv:2307.04721}, 2023.

\bibitem[Nguyen \& Grover(2022)Nguyen and Grover]{nguyen2022transformer}
Tung Nguyen and Aditya Grover.
\newblock Transformer neural processes: Uncertainty-aware meta learning via sequence modeling.
\newblock In \emph{ICML}, 2022.

\bibitem[Nguyen et~al.(2023)Nguyen, Agrawal, and Grover]{nguyen2023expt}
Tung Nguyen, Sudhanshu Agrawal, and Aditya Grover.
\newblock Expt: Synthetic pretraining for few-shot experimental design.
\newblock In \emph{NeurIPS}, 2023.

\bibitem[Nie et~al.(2023)Nie, Cheng, Kolobov, and Swaminathan]{nie2023importance}
Allen Nie, Ching-An Cheng, Andrey Kolobov, and Adith Swaminathan.
\newblock Importance of directional feedback for llm-based optimizers.
\newblock In \emph{NeurIPS 2023 Foundation Models for Decision Making Workshop}, 2023.

\bibitem[Olivecrona et~al.(2017)Olivecrona, Blaschke, Engkvist, and Chen]{olivecrona2017molecular}
Marcus Olivecrona, Thomas Blaschke, Ola Engkvist, and Hongming Chen.
\newblock Molecular de-novo design through deep reinforcement learning.
\newblock \emph{Journal of cheminformatics}, 9\penalty0 (1):\penalty0 1--14, 2017.

\bibitem[Pei et~al.()Pei, Zhang, Zhu, Wu, Gao, Wu, Xia, and Yan]{pei2023biot5}
Qizhi Pei, Wei Zhang, Jinhua Zhu, Kehan Wu, Kaiyuan Gao, Lijun Wu, Yingce Xia, and Rui Yan.
\newblock Biot5: Enriching cross-modal integration in biology with chemical knowledge and natural language associations.
\newblock In \emph{The 2023 Conference on Empirical Methods in Natural Language Processing}.

\bibitem[Raffel et~al.(2020)Raffel, Shazeer, Roberts, Lee, Narang, Matena, Zhou, Li, and Liu]{raffel2020exploring}
Colin Raffel, Noam Shazeer, Adam Roberts, Katherine Lee, Sharan Narang, Michael Matena, Yanqi Zhou, Wei Li, and Peter~J Liu.
\newblock Exploring the limits of transfer learning with a unified text-to-text transformer.
\newblock \emph{Journal of machine learning research}, 21\penalty0 (140):\penalty0 1--67, 2020.

\bibitem[Ramos et~al.(2023)Ramos, Michtavy, Porosoff, and White]{ramos2023bayesian}
Mayk~Caldas Ramos, Shane~S Michtavy, Marc~D Porosoff, and Andrew~D White.
\newblock Bayesian optimization of catalysts with in-context learning.
\newblock \emph{arXiv preprint arXiv:2304.05341}, 2023.

\bibitem[Rankovi{\'c} \& Schwaller(2023)Rankovi{\'c} and Schwaller]{rankovic2023bochemian}
Bojana Rankovi{\'c} and Philippe Schwaller.
\newblock Bochemian: Large language model embeddings for bayesian optimization of chemical reactions.
\newblock In \emph{NeurIPS 2023 Workshop on Adaptive Experimental Design and Active Learning in the Real World}, 2023.

\bibitem[Sarkisyan et~al.(2016)Sarkisyan, Bolotin, Meer, Usmanova, Mishin, Sharonov, Ivankov, Bozhanova, Baranov, Soylemez, et~al.]{sarkisyan2016local}
Karen~S Sarkisyan, Dmitry~A Bolotin, Margarita~V Meer, Dinara~R Usmanova, Alexander~S Mishin, George~V Sharonov, Dmitry~N Ivankov, Nina~G Bozhanova, Mikhail~S Baranov, Onuralp Soylemez, et~al.
\newblock Local fitness landscape of the green fluorescent protein.
\newblock \emph{Nature}, 533\penalty0 (7603):\penalty0 397--401, 2016.

\bibitem[Shen et~al.(2023)Shen, Li, Dery, Staten, Khodak, Neubig, and Talwalkar]{shen2023cross}
Junhong Shen, Liam Li, Lucio~M Dery, Corey Staten, Mikhail Khodak, Graham Neubig, and Ameet Talwalkar.
\newblock Cross-modal fine-tuning: Align then refine.
\newblock \emph{arXiv preprint arXiv:2302.05738}, 2023.

\bibitem[Sprueill et~al.(2024)Sprueill, Edwards, Agarwal, Olarte, Sanyal, Johnston, Liu, Ji, and Choudhury]{sprueill2024chemreasoner}
Henry~W Sprueill, Carl Edwards, Khushbu Agarwal, Mariefel~V Olarte, Udishnu Sanyal, Conrad Johnston, Hongbin Liu, Heng Ji, and Sutanay Choudhury.
\newblock Chemreasoner: Heuristic search over a large language model's knowledge space using quantum-chemical feedback.
\newblock \emph{arXiv preprint arXiv:2402.10980}, 2024.

\bibitem[Sterling \& Irwin(2015)Sterling and Irwin]{sterling2015zinc}
Teague Sterling and John~J Irwin.
\newblock Zinc 15--ligand discovery for everyone.
\newblock \emph{Journal of chemical information and modeling}, 55\penalty0 (11):\penalty0 2324--2337, 2015.

\bibitem[Team et~al.(2023)Team, Anil, Borgeaud, Wu, Alayrac, Yu, Soricut, Schalkwyk, Dai, Hauth, et~al.]{team2023gemini}
Gemini Team, Rohan Anil, Sebastian Borgeaud, Yonghui Wu, Jean-Baptiste Alayrac, Jiahui Yu, Radu Soricut, Johan Schalkwyk, Andrew~M Dai, Anja Hauth, et~al.
\newblock Gemini: a family of highly capable multimodal models.
\newblock \emph{arXiv preprint arXiv:2312.11805}, 2023.

\bibitem[Touvron et~al.(2023{\natexlab{a}})Touvron, Lavril, Izacard, Martinet, Lachaux, Lacroix, Rozi{\`e}re, Goyal, Hambro, Azhar, et~al.]{touvron2023llama}
Hugo Touvron, Thibaut Lavril, Gautier Izacard, Xavier Martinet, Marie-Anne Lachaux, Timoth{\'e}e Lacroix, Baptiste Rozi{\`e}re, Naman Goyal, Eric Hambro, Faisal Azhar, et~al.
\newblock Llama: Open and efficient foundation language models.
\newblock \emph{arXiv preprint arXiv:2302.13971}, 2023{\natexlab{a}}.

\bibitem[Touvron et~al.(2023{\natexlab{b}})Touvron, Martin, Stone, Albert, Almahairi, Babaei, Bashlykov, Batra, Bhargava, Bhosale, et~al.]{touvron2023llama2}
Hugo Touvron, Louis Martin, Kevin Stone, Peter Albert, Amjad Almahairi, Yasmine Babaei, Nikolay Bashlykov, Soumya Batra, Prajjwal Bhargava, Shruti Bhosale, et~al.
\newblock Llama 2: Open foundation and fine-tuned chat models.
\newblock \emph{arXiv preprint arXiv:2307.09288}, 2023{\natexlab{b}}.

\bibitem[Tripp et~al.(2021)Tripp, Simm, and Hern{\'a}ndez-Lobato]{tripp2021fresh}
Austin Tripp, Gregor~NC Simm, and Jos{\'e}~Miguel Hern{\'a}ndez-Lobato.
\newblock A fresh look at de novo molecular design benchmarks.
\newblock In \emph{NeurIPS 2021 AI for Science Workshop}, 2021.

\bibitem[Tsimpoukelli et~al.(2021)Tsimpoukelli, Menick, Cabi, Eslami, Vinyals, and Hill]{tsimpoukelli2021multimodal}
Maria Tsimpoukelli, Jacob~L Menick, Serkan Cabi, SM~Eslami, Oriol Vinyals, and Felix Hill.
\newblock Multimodal few-shot learning with frozen language models.
\newblock \emph{Advances in Neural Information Processing Systems}, 34:\penalty0 200--212, 2021.

\bibitem[V{\"o}lker et~al.(2024)V{\"o}lker, Rug, Jablonka, and Kruschwitz]{volker2024llms}
Christoph V{\"o}lker, Tehseen Rug, Kevin~Maik Jablonka, and Sabine Kruschwitz.
\newblock Llms can design sustainable concrete--a systematic benchmark.
\newblock 2024.

\bibitem[Wang et~al.(2024)Wang, Skreta, Ser, Gao, Kong, Strieth-Kalthoff, Duan, Zhuang, Yu, Zhu, et~al.]{wang2024efficient}
Haorui Wang, Marta Skreta, Cher-Tian Ser, Wenhao Gao, Lingkai Kong, Felix Strieth-Kalthoff, Chenru Duan, Yuchen Zhuang, Yue Yu, Yanqiao Zhu, et~al.
\newblock Efficient evolutionary search over chemical space with large language models.
\newblock \emph{arXiv preprint arXiv:2406.16976}, 2024.

\bibitem[Wei et~al.(2022)Wei, Wang, Schuurmans, Bosma, Xia, Chi, Le, Zhou, et~al.]{wei2022chain}
Jason Wei, Xuezhi Wang, Dale Schuurmans, Maarten Bosma, Fei Xia, Ed~Chi, Quoc~V Le, Denny Zhou, et~al.
\newblock Chain-of-thought prompting elicits reasoning in large language models.
\newblock \emph{Advances in Neural Information Processing Systems}, 35:\penalty0 24824--24837, 2022.

\bibitem[Weininger(1988)]{weininger1988smiles}
David Weininger.
\newblock Smiles, a chemical language and information system. 1. introduction to methodology and encoding rules.
\newblock \emph{Journal of chemical information and computer sciences}, 28\penalty0 (1):\penalty0 31--36, 1988.

\bibitem[Yang et~al.(2023)Yang, Wang, Lu, Liu, Le, Zhou, and Chen]{yang2023large}
Chengrun Yang, Xuezhi Wang, Yifeng Lu, Hanxiao Liu, Quoc~V Le, Denny Zhou, and Xinyun Chen.
\newblock Large language models as optimizers.
\newblock \emph{arXiv preprint arXiv:2309.03409}, 2023.

\bibitem[Ye et~al.(2023)Ye, Cai, Lai, Wang, Huang, Wang, Liu, and Zeng]{ye2023drugassist}
Geyan Ye, Xibao Cai, Houtim Lai, Xing Wang, Junhong Huang, Longyue Wang, Wei Liu, and Xiangxiang Zeng.
\newblock Drugassist: A large language model for molecule optimization.
\newblock \emph{arXiv preprint arXiv:2401.10334}, 2023.

\bibitem[Zhang et~al.(2023)Zhang, Desai, Bae, Lorraine, and Ba]{zhang2023using}
Michael~R Zhang, Nishkrit Desai, Juhan Bae, Jonathan Lorraine, and Jimmy Ba.
\newblock Using large language models for hyperparameter optimization.
\newblock \emph{arXiv e-prints}, pp.\  arXiv--2312, 2023.

\bibitem[Zoph \& Le(2016)Zoph and Le]{zoph2016neural}
Barret Zoph and Quoc~V Le.
\newblock Neural architecture search with reinforcement learning.
\newblock \emph{arXiv preprint arXiv:1611.01578}, 2016.

\end{thebibliography}
\bibliographystyle{iclr2025_conference}

\clearpage
\appendix
\section{\name{} implementation details} \label{sec:imp_details}
\subsection{Molecular intrinsic functions} \label{sec:intrinsic_props}
We utilize $47$ intrinsic properties of molecules for pretraining \name{}. Table~\ref{tab:intrinsic_props} shows the intrinsic properties and their explanation.

\begin{table}[h!]
\centering
\caption{Inherent Properties of Molecules and their Explanations}
\label{tab:intrinsic_props}
\resizebox{0.87\textwidth}{!}{%
\begin{tabular}{@{}ll@{}}
\toprule
\textbf{Property} & \textbf{Explanation} \\ \midrule
Molecular Weight & Total mass of all atoms in the molecule. \\
Number of Rotatable Bonds & Bonds that allow free rotation around themselves. \\
Number of Rings & Count of ring structures in the molecule. \\
Number of H Donors & Atoms in the molecule that can donate a hydrogen atom. \\
Number of H Acceptors & Atoms in the molecule capable of accepting a hydrogen atom. \\
Num Aromatic Rings & Count of rings with a pattern of alternating single and double bonds. \\
Num Aliphatic Rings & Count of non-aromatic rings in the molecule. \\
Num Saturated Rings & Rings with single bonds only. \\
Num Heteroatoms & Atoms other than carbon or hydrogen. \\
Fraction Csp3 & Fraction of carbon atoms bonded with a single pair of electrons. \\
Heavy Atom Count & Count of all atoms except hydrogen. \\
Num Valence Electrons & Total number of electrons that can participate in the formation of chemical bonds. \\
Num Aromatic CarboRings & Aromatic rings composed solely of carbon atoms. \\
Num Aromatic HeteroRings & Aromatic rings containing at least one heteroatom. \\
Num Saturated CarboRings & Saturated rings made only of carbon atoms. \\
Num Saturated HeteroRings & Saturated rings containing at least one heteroatom. \\
BalabanJ & Topological index to quantify molecular branching. \\
BertzCT & A measure of structural complexity of the molecule. \\
Ipc & Information content on the vertex degree. \\
HallKierAlpha & Valence connectivity index used in molecular shape analysis. \\
Kappa1 & Shape descriptor based on the skeleton of the molecule. \\
Kappa2 & Hydrogen suppressed graph descriptor. \\
Kappa3 & Hydrogen complete graph descriptor. \\
Chi0 & Randić molecular connectivity index. \\
Chi1 & Valence modified Randić molecular connectivity index. \\
Chi0n & Randić connectivity index normalized. \\
Chi1n & Valence modified Randić connectivity index normalized. \\
Chi2n & Second order Randić connectivity index normalized. \\
Chi3n & Third order Randić connectivity index normalized. \\
Chi4n & Fourth order Randić connectivity index normalized. \\
Chi0v & Randić connectivity index for valence electrons. \\
Chi1v & First order valence molecular connectivity index. \\
Chi2v & Second order valence molecular connectivity index. \\
Chi3v & Third order valence molecular connectivity index. \\
Chi4v & Fourth order valence molecular connectivity index. \\
Molar Refractivity & Measure of the molecule's polarizability. \\
AMW & Average molecular weight of all atoms in the molecule. \\
Max Partial Charge & Maximum partial charge in the molecule. \\
Min Partial Charge & Minimum partial charge in the molecule. \\
Max Abs Partial Charge & Maximum absolute value of the partial charges in the molecule. \\
Min Abs Partial Charge & Minimum absolute value of the partial charges in the molecule. \\
Labute ASA & Labute's Approximate Surface Area, an estimate of the molecular surface area. \\
Max EState Index & Maximum electrotopological state index of the atoms in the molecule. \\
Min EState Index & Minimum electrotopological state index of the atoms in the molecule. \\
Max Abs EState Index & Maximum absolute value of the electrotopological state indices in the molecule. \\
Min Abs EState Index & Minimum absolute value of the electrotopological state indices in the molecule. \\
fr\_C\_O & Frequency of carbon-oxygen bonds in the molecule. \\ \bottomrule
\end{tabular}%
}
\end{table}

\subsection{Training details}
The $x$ embedding layer, $y$ embedding layer, and prediction layer in \name{} are MLPs with a hidden dimension of $1024$. We train \name{} for $20000$ steps with a batch size of $4$. For each data point in the batch, we randomly decide whether to sample an intrinsic or a synthetic function, with the probability of choosing synthetic functions being $0.1$. Each data point is a sequence of $(x,y)$ pairs with length $n \sim \mathcal{U}[64,800]$. If the function is an intrinsic function, we uniformly sample a property from Table~\ref{tab:intrinsic_props}, otherwise sample synthetic data following Equation~\eqref{eq:synthetic_sampling}.

We use Llama-2-7b~\citep{touvron2023llama2} as the base LLM, and use LoRA~\citep{hu2021lora} for parameter-efficient finetuning. We use a base learning rate of $5e-4$ with a linear warmup for $1000$ steps and a cosine decay for the remaining $19000$ steps. We use LoRA with a rank of $16$ and $\alpha$ scale of $16$.

\subsection{Black-box optimization hyperparameters} \label{app:bbo_hps}

We use Algorithm~\ref{alg:optimization} to optimize a black-box function with \name{}. We initialize the observed dataset $\mathcal{D}_{\text{obs}}$ with a population of $34$ molecules sampled randomly from ZINC. At each iteration, we use the best $34$ candidates in $\mathcal{D}_{\text{obs}}$ to generate new candidates via crossover and mutation operations, with the mutation rate being $0.01$. The candidate pool size $C$ is $100$. We predict the mean $\mu_i$ and standard deviation $\sigma_i$ for each candidate $x_i$ in the pool using \name{}. We employ a UCB acquisition function to compute the utility score $u_i = \mu_i + \beta\sigma_i$, which balances exploration and exploitation. Following~\citep{gao2022sample}, we set $\beta = 10^b$, where $b \sim \mathcal{U}[-0.5, 1.5]$. We then pick $k=15$ candidates with the highest utility scores. We evaluate each selected candidate $x_j$ using the black-box function $f$, and add the new data point $(x_j, y_j)$ to the observed dataset $\mathcal{D}_{\text{obs}}$. The process continues with the updated observed dataset, and stops when $|\mathcal{D}_{\text{obs}}| = 1000.$

When predicting $\mu_i, \sigma_i = f_\theta(x_i, \mathcal{D}_{\text{obs}})$, we normalize all the $y's$ values in $\mathcal{D}_{\text{obs}}$ to have mean $0$ and standard deviation $1$. This is to resemble the finetuning data distribution of \name{}. We then denormalize $\mu_i$ and $\sigma_i$ to the original space.

\subsection{Black-box optimization with \name{}}
Algorithm~\ref{alg:optimization} outlines the optimization algorithm using \name{} as the surrogate model.
\begin{algorithm}[h]
\caption{Black-box optimization with \name{}
}
\label{alg:optimization}
\begin{algorithmic}
\small
\REQUIRE objective $f$, \name{} model $f_\theta$, budget $B$, candidate pool size $C$, acquisition function $\alpha$, batch size $k$
\STATE Initialize $\mathcal{D}_{\text{obs}} = \{\}$
\WHILE{$|\mathcal{D}_{\text{obs}}| < B$}
    \STATE Generate a set of candidates $\{x_i\}_{i=1}^C$
    \FOR{each candidate $x_i$}
        \STATE Predict $\mu_i, \sigma_i = f_\theta(x_i, \mathcal{D}_{\text{obs}})$
        \STATE Compute utility score $u_i = \alpha(\mu_i, \sigma_i)$
    \ENDFOR
    \STATE Select $k$ candidates with the highest utility scores
    \FOR{each selected candidate $x_j$}
        \STATE Evaluate $x_j$ using the actual objective $y_{j} = f(x_j)$
        \STATE Add $(x_j, y_j)$ to the observation dataset $\mathcal{D}_{\text{obs}}$
    \ENDFOR
\ENDWHILE
\end{algorithmic}
\end{algorithm}

\section{Baseline details}

\paragraph{TNP} is a transformer-based architecture for in-context learning. We refer to~\citet{nguyen2022transformer} for more details about TNP. We train a TNP model with $16$ attention layers and $2048$ hidden dimensions. Other hyperparameters are the same as for \name{}. After training, we use TNP for black-box optimization using Algorithm~\ref{alg:optimization} with the same optimization hyperparameters but replace \name{} with TNP.

\paragraph{GP BO} replaces the \name{} surrogate model in Algorithm~\ref{alg:optimization} with a Gaussian Process with a Tanimoto kernel. We optimize the Gaussian Process hyperparameters via maximum likelihood estimation on the initial population sampled from ZINC. 

\paragraph{Graph GA} is a model-free variant of Algorithm~\ref{alg:optimization}. Specifically, at each iteration, Graph GA generates a set of candidates using the same crossover and mutation operations, and directly evaluates and adds them to $\mathcal{D}_{\text{obs}}$, since it does not employ a surrogate model.

\paragraph{REINVENT} adopts a policy-based RL approach to finetune a pretrained RNN to generate SMILES strings with high returns. At each optimization iteration, we sample a set of molecules from the finetuned RNN, evaluate these molecules using the black-box function $f$, and add the new data points to $\mathcal{D}_{\text{obs}}$. We refer to~\citet{gao2022sample} for more details of the algorithm and other hyperparameters.

\subsection{Hyperparameter tuning for the baselines} \label{app:hp_tuning}

To ensure the baselines achieve the best possible performance for the new PMO-1K benchmark, we performed extensive hyperparameter tuning for each baseline on the first $5$ tasks with grid search, and used the optimal hyperparameters for the rest of the tasks. Table~\ref{tab:hp_tuning} specifies the grid search for each method.

\begin{table}[h!]
\centering
\caption{Grid search and optimal hyperparameters for the baseline methods.}
\label{tab:hp_tuning}
\resizebox{\textwidth}{!}{%
\begin{tabular}{@{}l p{7cm} p{7cm}@{}}
\toprule
\textbf{Method} & \textbf{Grid Search} & \textbf{Optimal Hyperparameters} \\ \midrule
\textbf{GP BO} & 
\texttt{population\_size $\in$ \{50, 100, 150, ..., 350\}} \newline
\texttt{offspring\_size $\in$ \{50, 100, 150\}} \newline
\texttt{kept\_offspring\_size $\in$ \{5, 10, ..., 50\}} & 
\texttt{population\_size = 50} \newline
\texttt{offspring\_size = 100} \newline
\texttt{kept\_offspring\_size = 15} \\ \midrule
\textbf{Graph GA} & 
\texttt{population\_size $\in$ \{50, 100, 150, ..., 350\}} \newline
\texttt{offspring\_size $\in$ \{50, 100, 150\}} & 
\texttt{population\_size = 50} \newline
\texttt{offspring\_size = 100} \\ \midrule
\textbf{REINVENT} & 
\texttt{batch\_size $\in$ \{4, 8, 16, 32, 64\}} \newline
\texttt{experience\_replay $\in$ \{4, 8, 16, 24, 32\}} & 
\texttt{batch\_size = 16} \newline
\texttt{experience\_replay = 24} \\ \midrule
\textbf{Genetic GFN} & 
\texttt{learning\_rate $\in$ \{0.0001, 0.0005\}} \newline
\texttt{batch\_size $\in$ \{4, 8, 16, 32, 64\}} \newline
\texttt{num\_keep $\in$ \{128, 256, 512, 1024\}} \newline
\texttt{offspring\_size $\in$ \{2, 4, 8\}} \newline
\texttt{ga\_generations $\in$ \{1, 2\}} & 
\texttt{learning\_rate = 0.0001} \newline
\texttt{batch\_size = 8} \newline
\texttt{num\_keep = 128} \newline
\texttt{offspring\_size = 2} \newline
\texttt{ga\_generations = 1} \\ \midrule
\textbf{Augmented Memory} & 
\texttt{batch\_size $\in$ \{4, 8, 16, 32, 64\}} \newline
\texttt{replay\_buffer\_size $\in$ \{50, 100, 150\}} & 
\texttt{batch\_size = 32} \newline
\texttt{replay\_buffer\_size = 100} \\ \bottomrule
\end{tabular}%
}
\end{table}

\section{Additional results}
\subsection{Additional metrics}
In addition to AUC Average Top-10, we measure the optimization performance of different methods on AUC Average Top-1 and AUC Average Top-100 for a more comprehensive comparison. Table~\ref{tab:auc_top1} and~\ref{tab:auc_top100} show AUC Average Top-1 and AUC Average Top-100 performances, respectively.

\begin{table*}[h!]
    \centering
    \caption{The performance of \name{} and the baselines on $21$ optimization tasks in PMO with AUC Average Top-1 metric. A higher score is better. We report the mean and stddev of scores averaged over $5$ random seeds. We use \textcolor{blue}{\textbf{blue}} and \textcolor{violet}{\textbf{violet}} to denote the best and second-best method for each task.}
    \label{tab:auc_top1}
    \resizebox{0.95\linewidth}{!}{
    \begin{tabular}{cccccc}
        \toprule
        Task & GP BO & Graph GA & \name{} & REINVENT & TNP \\
        \midrule
        \texttt{albuterol\_similarity} & $\violet{0.672 \pm 0.109}$ & $0.647 \pm 0.080$ & $\blue{0.695 \pm 0.150}$ & $0.572 \pm 0.026$ & $0.611 \pm 0.042$ \\
        \texttt{amlodipine\_mpo} & $\violet{0.538 \pm 0.016}$ & $0.526 \pm 0.017$ & $\blue{0.560 \pm 0.026}$ & $0.500 \pm 0.016$ & $0.513 \pm 0.016$ \\
        \texttt{celecoxib\_rediscovery} & $0.434 \pm 0.052$ & $0.466 \pm 0.062$ & $\blue{0.492 \pm 0.079}$ & $0.415 \pm 0.031$ & $\violet{0.482 \pm 0.067}$ \\
        \texttt{deco\_hop} & $\violet{0.598 \pm 0.013}$ & $0.590 \pm 0.005$ & $\blue{0.603 \pm 0.012}$ & $0.585 \pm 0.010$ & $0.597 \pm 0.002$ \\
        \texttt{drd2\_current} & $0.895 \pm 0.067$ & $\violet{0.898 \pm 0.048}$ & $\blue{0.902 \pm 0.055}$ & $0.867 \pm 0.077$ & $0.831 \pm 0.043$ \\
        \texttt{fexofenadine\_mpo} & $\blue{0.728 \pm 0.022}$ & $0.691 \pm 0.011$ & $\violet{0.719 \pm 0.025}$ & $0.696 \pm 0.012$ & $0.706 \pm 0.014$ \\
        \texttt{isomers\_c7h8n2o2} & $0.576 \pm 0.154$ & $0.815 \pm 0.120$ & $\violet{0.834 \pm 0.109}$ & $\blue{0.846 \pm 0.070}$ & $0.761 \pm 0.145$ \\
        \texttt{isomers\_c9h10n2o2pf2cl} & $0.644 \pm 0.053$ & $0.708 \pm 0.083$ & $\violet{0.714 \pm 0.084}$ & $\blue{0.724 \pm 0.043}$ & $0.701 \pm 0.086$ \\
        \texttt{median1} & $0.235 \pm 0.016$ & $0.233 \pm 0.018$ & $\blue{0.242 \pm 0.020}$ & $0.229 \pm 0.015$ & $\violet{0.238 \pm 0.015}$ \\
        \texttt{median2} & $\blue{0.212 \pm 0.010}$ & $0.193 \pm 0.011$ & $0.201 \pm 0.009$ & $\violet{0.209 \pm 0.013}$ & $0.200 \pm 0.018$ \\
        \texttt{mestranol\_similarity} & $\blue{0.449 \pm 0.028}$ & $0.387 \pm 0.020$ & $\violet{0.445 \pm 0.014}$ & $0.433 \pm 0.034$ & $0.406 \pm 0.011$ \\
        \texttt{osimertinib\_mpo} & $\blue{0.788 \pm 0.008}$ & $0.777 \pm 0.008$ & $\violet{0.781 \pm 0.007}$ & $0.780 \pm 0.009$ & $0.776 \pm 0.007$ \\
        \texttt{perindopril\_mpo} & $\violet{0.475 \pm 0.019}$ & $0.460 \pm 0.025$ & $\blue{0.492 \pm 0.011}$ & $0.432 \pm 0.010$ & $0.457 \pm 0.012$ \\
        \texttt{qed} & $0.926 \pm 0.011$ & $0.930 \pm 0.004$ & $\blue{0.935 \pm 0.002}$ & $\violet{0.934 \pm 0.003}$ & $0.931 \pm 0.001$ \\
        \texttt{ranolazine\_mpo} & $\blue{0.729 \pm 0.024}$ & $0.684 \pm 0.015$ & $\violet{0.711 \pm 0.028}$ & $0.657 \pm 0.048$ & $0.669 \pm 0.032$ \\
        \texttt{scaffold\_hop} & $\violet{0.486 \pm 0.010}$ & $0.475 \pm 0.008$ & $\blue{0.491 \pm 0.013}$ & $0.468 \pm 0.010$ & $0.484 \pm 0.019$ \\
        \texttt{sitagliptin\_mpo} & $0.268 \pm 0.098$ & $0.281 \pm 0.069$ & $\blue{0.363 \pm 0.114}$ & $\violet{0.333 \pm 0.030}$ & $0.274 \pm 0.044$ \\
        \texttt{thiothixene\_rediscovery} & $\blue{0.371 \pm 0.046}$ & $0.351 \pm 0.029$ & $\violet{0.368 \pm 0.041}$ & $0.345 \pm 0.026$ & $0.332 \pm 0.041$ \\
        \texttt{troglitazone\_rediscovery} & $\blue{0.329 \pm 0.019}$ & $0.289 \pm 0.021$ & $\violet{0.309 \pm 0.033}$ & $0.276 \pm 0.009$ & $0.286 \pm 0.012$ \\
        \texttt{valsartan\_smarts} & $\violet{0.000 \pm 0.000}$ & $0.000 \pm 0.000$ & $0.000 \pm 0.000$ & $\blue{0.000 \pm 0.000}$ & $0.000 \pm 0.000$ \\
        \texttt{zaleplon\_mpo} & $0.431 \pm 0.031$ & $0.418 \pm 0.022$ & $\violet{0.435 \pm 0.027}$ & $\blue{0.456 \pm 0.020}$ & $0.428 \pm 0.022$ \\
        \midrule
        Sum of scores ($\uparrow$) & $10.784$ & $\violet{10.818}$ & $\blue{11.291}$ & $10.755$ & $10.683$ \\
        Mean rank ($\downarrow$) & $\violet{2.52}$ & $3.57$ & $\blue{1.62}$ & $3.48$ & $3.75$ \\
        \bottomrule
    \end{tabular}
    }
\end{table*}

\begin{table*}[h!]
    \centering
    \caption{The performance of \name{} and the baselines on $21$ optimization tasks in PMO with AUC Average Top-100 metric. A higher score is better. We report the mean and stddev of scores averaged over $5$ random seeds. We use \textcolor{blue}{\textbf{blue}} and \textcolor{violet}{\textbf{violet}} to denote the best and second-best method for each task.}
    \label{tab:auc_top100}
    \resizebox{0.95\linewidth}{!}{
    \begin{tabular}{cccccc}
        \toprule
        Task & GP BO & Graph GA & \name{} & REINVENT & TNP \\
        \midrule
        \texttt{albuterol\_similarity} & $\violet{0.548 \pm 0.100}$ & $0.470 \pm 0.042$ & $\blue{0.563 \pm 0.093}$ & $0.395 \pm 0.012$ & $0.448 \pm 0.028$ \\
        \texttt{amlodipine\_mpo} & $\violet{0.458 \pm 0.008}$ & $0.422 \pm 0.014$ & $\blue{0.486 \pm 0.025}$ & $0.407 \pm 0.005$ & $0.420 \pm 0.013$ \\
        \texttt{celecoxib\_rediscovery} & $\violet{0.363 \pm 0.040}$ & $0.346 \pm 0.036$ & $\blue{0.372 \pm 0.070}$ & $0.296 \pm 0.024$ & $0.346 \pm 0.026$ \\
        \texttt{deco\_hop} & $\violet{0.579 \pm 0.013}$ & $0.563 \pm 0.006$ & $\blue{0.583 \pm 0.009}$ & $0.550 \pm 0.006$ & $0.568 \pm 0.004$ \\
        \texttt{drd2\_current} & $\blue{0.741 \pm 0.097}$ & $0.605 \pm 0.086$ & $\violet{0.725 \pm 0.092}$ & $0.615 \pm 0.098$ & $0.556 \pm 0.095$ \\
        \texttt{fexofenadine\_mpo} & $\blue{0.645 \pm 0.018}$ & $0.588 \pm 0.008$ & $\violet{0.636 \pm 0.022}$ & $0.549 \pm 0.004$ & $0.599 \pm 0.016$ \\
        \texttt{isomers\_c7h8n2o2} & $0.300 \pm 0.142$ & $\blue{0.535 \pm 0.091}$ & $0.450 \pm 0.149$ & $\violet{0.511 \pm 0.058}$ & $0.492 \pm 0.115$ \\
        \texttt{isomers\_c9h10n2o2pf2cl} & $\violet{0.474 \pm 0.038}$ & $0.441 \pm 0.068$ & $\blue{0.535 \pm 0.067}$ & $0.445 \pm 0.027$ & $0.447 \pm 0.049$ \\
        \texttt{median1} & $\blue{0.175 \pm 0.022}$ & $0.168 \pm 0.013$ & $0.166 \pm 0.019$ & $0.162 \pm 0.007$ & $\violet{0.170 \pm 0.008}$ \\
        \texttt{median2} & $\blue{0.184 \pm 0.006}$ & $0.158 \pm 0.008$ & $\violet{0.175 \pm 0.010}$ & $0.155 \pm 0.006$ & $0.162 \pm 0.009$ \\
        \texttt{mestranol\_similarity} & $\blue{0.379 \pm 0.020}$ & $0.311 \pm 0.016$ & $\violet{0.361 \pm 0.030}$ & $0.302 \pm 0.016$ & $0.314 \pm 0.003$ \\
        \texttt{osimertinib\_mpo} & $\blue{0.706 \pm 0.006}$ & $0.667 \pm 0.008$ & $\violet{0.694 \pm 0.010}$ & $0.623 \pm 0.014$ & $0.671 \pm 0.006$ \\
        \texttt{perindopril\_mpo} & $\violet{0.405 \pm 0.019}$ & $0.357 \pm 0.012$ & $\blue{0.424 \pm 0.007}$ & $0.332 \pm 0.011$ & $0.359 \pm 0.010$ \\
        \texttt{qed} & $0.853 \pm 0.010$ & $0.854 \pm 0.011$ & $\blue{0.882 \pm 0.007}$ & $\violet{0.874 \pm 0.003}$ & $0.857 \pm 0.003$ \\
        \texttt{ranolazine\_mpo} & $\blue{0.633 \pm 0.020}$ & $0.462 \pm 0.022$ & $\violet{0.617 \pm 0.021}$ & $0.436 \pm 0.040$ & $0.468 \pm 0.042$ \\
        \texttt{scaffold\_hop} & $\violet{0.462 \pm 0.006}$ & $0.435 \pm 0.008$ & $\blue{0.462 \pm 0.006}$ & $0.415 \pm 0.009$ & $0.440 \pm 0.010$ \\
        \texttt{sitagliptin\_mpo} & $0.133 \pm 0.062$ & $0.103 \pm 0.032$ & $\blue{0.171 \pm 0.045}$ & $\violet{0.134 \pm 0.016}$ & $0.100 \pm 0.023$ \\
        \texttt{thiothixene\_rediscovery} & $\blue{0.311 \pm 0.030}$ & $0.270 \pm 0.015$ & $\violet{0.299 \pm 0.026}$ & $0.256 \pm 0.015$ & $0.261 \pm 0.024$ \\
        \texttt{troglitazone\_rediscovery} & $\blue{0.283 \pm 0.014}$ & $0.228 \pm 0.008$ & $\violet{0.258 \pm 0.024}$ & $0.201 \pm 0.008$ & $0.230 \pm 0.005$ \\
        \texttt{valsartan\_smarts} & $\violet{0.000 \pm 0.000}$ & $0.000 \pm 0.000$ & $0.000 \pm 0.000$ & $\blue{0.000 \pm 0.000}$ & $0.000 \pm 0.000$ \\
        \texttt{zaleplon\_mpo} & $\violet{0.301 \pm 0.036}$ & $0.258 \pm 0.016$ & $\blue{0.318 \pm 0.018}$ & $0.296 \pm 0.009$ & $0.257 \pm 0.013$ \\
        \midrule
        Sum of scores ($\uparrow$) & $\violet{8.933}$ & $8.242$ & $\blue{9.175}$ & $7.954$ & $8.167$ \\
        Mean rank ($\downarrow$) & $\violet{1.95}$ & $3.62$ & $\blue{1.76}$ & $4.14$ & $3.45$ \\
        \bottomrule
    \end{tabular}
    }
\end{table*}

\clearpage
\subsection{LICO with LLMs trained on molecular corpora}
\begin{figure}[h]
    \centering
    \includegraphics[width=0.8\linewidth]{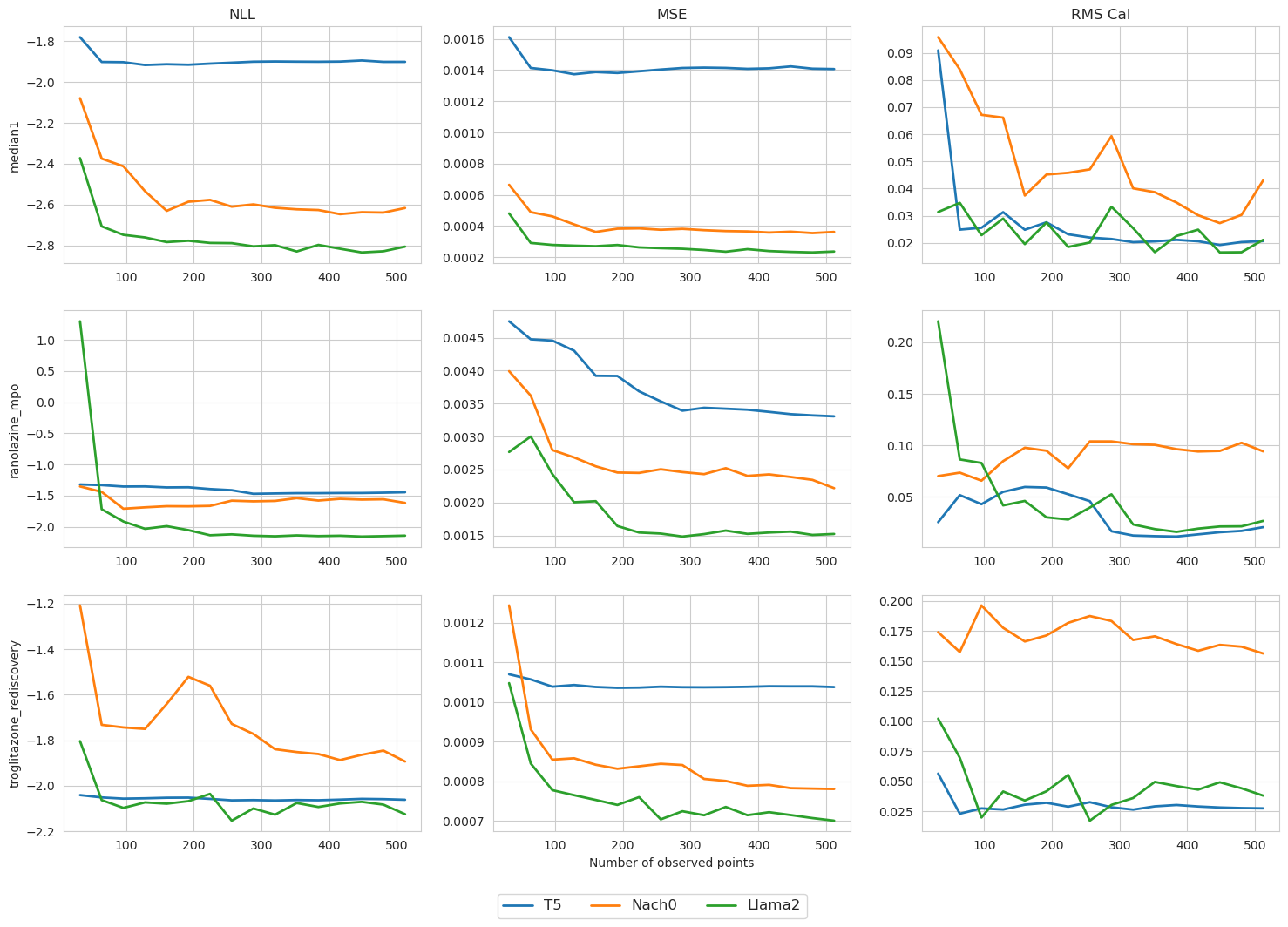}
    \caption{Predictive performance of LICO with T5-base, Nach0, and Llama-2 as the backbones.}
    \label{fig:molecule_llm_prediction}
\end{figure}

One may wonder whether using an LLM finetuned on molecular corpora helps improve the performance of LICO. To answer this question, we compare the performance of LICO with two different base LLMs: T5-base~\citep{raffel2020exploring} and Nach0-base~\citep{livne2024nach0}, which finetunes T5-base on molecule corpora. Due to time constraints, we did not perform optimization with these new models, but compared their predictive performance instead. Figure~\ref{fig:molecule_llm_prediction} summarizes the results. There are two interesting observations from this figure. First, Nach0 works significantly better than plain T5, confirming the hypothesis that proper finetuning of a language model on molecule data helps boost its in-context property prediction in LICO.
Second, Llama-2 works better than Nach0. As we explained in the paper, because we perform in-context learning in the embedding space of the language model, we rely on the general pattern-matching capability of the model, i.e., the ability to extract the relationship between embeddings of x and y from examples. From this perspective, it is not surprising that Llama-2 works better than a T5-based model, since it is much larger and has been pretrained with a lot more data, leading to a superior pattern-matching capability. One more benefit of using general LLMs like Llama is that they are domain-agnostic, which means we can finetune them for other non-molecule domains as well.

\subsection{LICO vs GPT-4 for molecular property prediction}
One may wonder if we can prompt state-of-the-art LLMs like GPT-4 to perform molecular property prediction. To investigate this, we conducted an experiment where we prompted GPT-4o to perform in-context property prediction in the text space. The following text box shows the prompt we used.

\begin{tcolorbox}[
    colback=blue!5,  %
    colframe=blue!75,  %
    boxrule=0.5pt,
    arc=4pt
]
I will give you a list of molecules and their corresponding property values.
Based on these examples, your task is to predict the property value of a new molecule.
Please provide your answer as a single number placed inside a pair of parentheses {} without any other information.
For example, if you think the property value of the new molecule is 0.5, you should write {0.5}.

Molecule: [m\textsubscript{1}], Property: [p\textsubscript{1}]

Molecule: [m\textsubscript{2}], Property: [p\textsubscript{2}]

\dots

Molecule: [m\textsubscript{n}], Property:

\end{tcolorbox}

\begin{figure}[h]
    \centering
    \includegraphics[width=1.0\linewidth]{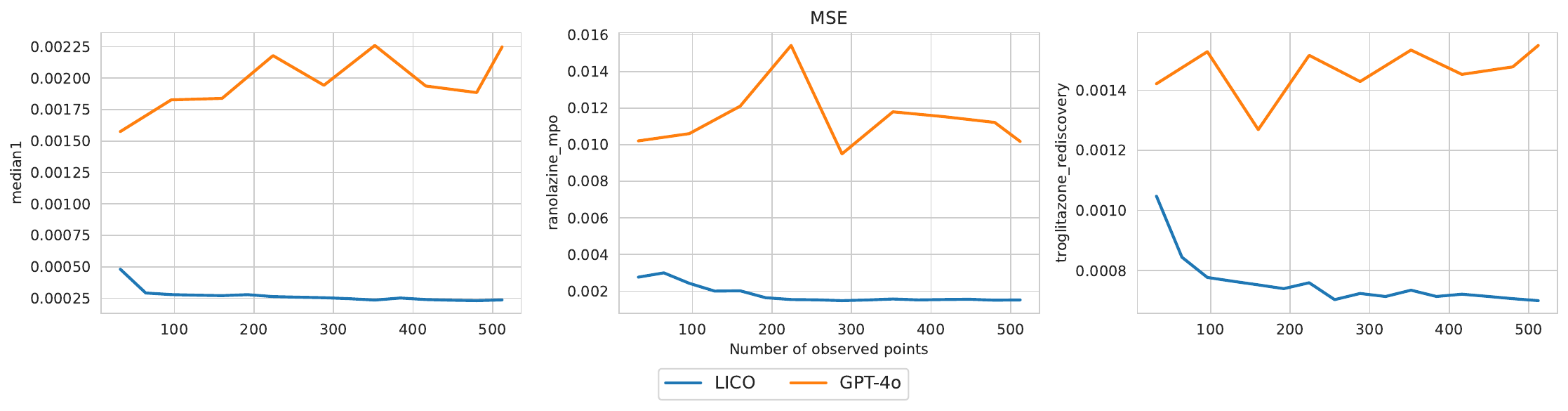}
    \caption{Predictive performance of LICO vs GPT-4o.}
    \label{fig:lico_vs_gpt}
\end{figure}

Figure~\ref{fig:lico_vs_gpt} compares the predictive performance of \name{} with GPT-4o on 3 tasks – \texttt{median1}, \texttt{ranolazine\_mpo}, and \texttt{troglitazone\_rediscovery}, similar to the paper. We vary the context length from 32 to 512, and for each context length average the mean squared error across 128 target molecules. The result shows that prompting GPT4-o directly in the text space performs poorly, while \name{} works much better and its performance improves as we increase the context length.

\subsection{LLM Embeddings for molecular optimization}
This work explored using LLMs as the in-context backbone for surrogate modeling in molecular optimization. Another way to make use of a pretrained LLM is to use the embeddings of an LLM as feature vectors for molecules and train a regressor on top of the embedding space. This section explores this idea with two regression models: a 4-layer MLP and a Gaussian Process (GP), both using the last hidden layer of the frozen LLama-2-7B model as the LLM embeddings. For the MLP baseline, we train the model on the full data buffer for 10 epochs with a batch size of $128$ and a learning rate of $1e-4$, for every 150 molecules collected (every time after each new set of molecules is selected and labeled by the oracle). For the GP baseline, we keep the same hyperparameters as the GP BO method that uses fingerprint features. We compare LICO with these two baselines in the PMO benchmark.

\begin{table*}[h!]
    \centering
    \caption{Comparison of LICO and other baselines with LLM embeddings on $23$ optimization tasks in PMO. A higher score is better. We report the mean and stddev of scores averaged over random seeds. We \textbf{bold} the best method for each task.}
    \label{tab:llm_embeddings}
    \resizebox{1.0\linewidth}{!}{
\begin{tabular}{cccc}
\toprule
Task & \name{} & MLP w/ LLM Embedding & GP BO w/ LLM Embedding \\
\midrule
\texttt{albuterol\_similarity} & $\bold{0.885 \pm 0.019}$ & $0.740 \pm 0.149$ & $0.868 \pm 0.175$ \\
\texttt{amlodipine\_mpo} & $\bold{0.679 \pm 0.027}$ & $0.621 \pm 0.058$ & $0.558 \pm 0.025$ \\
\texttt{celecoxib\_rediscovery} & $\bold{0.664 \pm 0.122}$ & $0.549 \pm 0.113$ & $0.633 \pm 0.187$ \\
\texttt{deco\_hop} & $\bold{0.619 \pm 0.015}$ & $0.594 \pm 0.006$ & $0.611 \pm 0.007$ \\
\texttt{drd2} & $0.928 \pm 0.018$ & $\bold{0.931 \pm 0.194}$ & $0.852 \pm 0.263$ \\
\texttt{fexofenadine\_mpo} & $\bold{0.772 \pm 0.023}$ & $0.734 \pm 0.045$ & $0.709 \pm 0.053$ \\
\texttt{gsk3b} & $\bold{0.876 \pm 0.045}$ & $0.793 \pm 0.034$ & $0.845 \pm 0.067$ \\
\texttt{isomers\_c7h8n2o2} & $0.939 \pm 0.022$ & $\bold{0.941 \pm 0.116}$ & $0.908 \pm 0.170$ \\
\texttt{isomers\_c9h10n2o2pf2cl} & $0.819 \pm 0.039$ & $\bold{0.835 \pm 0.111}$ & $0.739 \pm 0.119$ \\
\texttt{jnk3} & $\bold{0.731 \pm 0.037}$ & $0.725 \pm 0.021$ & $0.728 \pm 0.029$ \\
\texttt{median1} & $0.291 \pm 0.016$ & $0.282 \pm 0.039$ & $\bold{0.306 \pm 0.044}$ \\
\texttt{median2} & $0.280 \pm 0.019$ & $0.223 \pm 0.026$ & $0.280 \pm 0.037$ \\
\texttt{mestranol\_similarity} & $0.614 \pm 0.064$ & $\bold{0.770 \pm 0.120}$ & $0.612 \pm 0.150$ \\
\texttt{osimertinib\_mpo} & $\bold{0.820 \pm 0.012}$ & $0.809 \pm 0.011$ & $0.785 \pm 0.012$ \\
\texttt{perindopril\_mpo} & $\bold{0.557 \pm 0.028}$ & $0.497 \pm 0.024$ & $0.484 \pm 0.034$ \\
\texttt{qed} & $0.936 \pm 0.001$ & $\bold{0.946 \pm 0.002}$ & $0.946 \pm 0.002$ \\
\texttt{ranolazine\_mpo} & $\bold{0.774 \pm 0.008}$ & $0.736 \pm 0.085$ & $0.715 \pm 0.109$ \\
\texttt{scaffold\_hop} & $\bold{0.547 \pm 0.026}$ & $0.478 \pm 0.012$ & $0.506 \pm 0.010$ \\
\texttt{sitagliptin\_mpo} & $\bold{0.567 \pm 0.034}$ & $0.544 \pm 0.116$ & $0.429 \pm 0.129$ \\
\texttt{thiothixene\_rediscovery} & $0.514 \pm 0.037$ & $0.430 \pm 0.083$ & $\bold{0.521 \pm 0.125}$ \\
\texttt{troglitazone\_rediscovery} & $\bold{0.380 \pm 0.026}$ & $0.299 \pm 0.040$ & $0.370 \pm 0.083$ \\
\texttt{valsartan\_smarts} & $0.000 \pm 0.000$ & $0.000 \pm 0.000$ & $0.000 \pm 0.000$ \\
\texttt{zaleplon\_mpo} & $\bold{0.515 \pm 0.017}$ & $0.500 \pm 0.037$ & $0.483 \pm 0.043$ \\
\midrule
Sum of scores ($\uparrow$) & $\bold{14.708}$ & $13.975$ & $13.887$ \\
\bottomrule
\end{tabular}
}
\end{table*}

Table~\ref{tab:llm_embeddings} shows the superior performance of LICO against the two baselines, achieving the best performance in 14/23 tasks in PMO. In addition to the stronger empirical performance, a significant advantage of \name{} is the ability to generalize to any objective function via in-context learning without finetuning.

\section{Broader impact} \label{sec:broader}
Our work studies the application of large language models to black-box optimization, particularly in the domain of molecular optimization. This intersection of machine learning and optimization holds significant promise for advancing our understanding of LLMs' capabilities and limitations, and has significant potential in areas like material science and drug discovery. Our main goal is to enhance machine learning and optimization techniques, but it's also important to consider how these advancements might affect society, such as speeding up the development of new medicines and materials. 

\section{Compute Resources} \label{sec:compute}

All experiments in this paper are run on a cluster of $4$ A6000 GPUs, each with 49GB of memory.

\end{document}